\pdfoutput=1

\documentclass[11pt]{article}

\usepackage[preprint]{acl}
\usepackage{booktabs}

\usepackage{times}
\usepackage{latexsym}
\usepackage{enumitem}

\usepackage[T1]{fontenc}

\usepackage[utf8]{inputenc}

\usepackage{microtype}

\usepackage{inconsolata}

\usepackage{graphicx}

\usepackage{array}
\usepackage{multirow}
\usepackage{amsmath}  
\usepackage{amssymb}  
\usepackage{amsfonts} 
\usepackage{pifont}
\usepackage{subcaption}

\title{Unlocking the Power of LLM Uncertainty for \\Active In-Context Example Selection}

\author{Hsiu-Yuan Huang$^{1,2}$,
        Zichen Wu$^{1,2}$, 
        Yutong Yang$^{1,2}$,
        Junzhao Zhang$^{1,3}$,
        Yunfang Wu$^{1,2}$\thanks{~~Corresponding author.} \\
    $^{1}$National Key Laboratory for Multimedia Information Processing, Peking University \\ 
    $^{2}$School of Computer Science, Peking University, Beijing, China \\
    $^{3}$School of Software and Microelectronics, Peking University, Beijing, China \\
    \texttt{\{huang.hsiuyuan, zhangjunzhao\}@stu.pku.edu.cn},
    \texttt{\{wuzichen, yytpku, wuyf\}@pku.edu.cn},
    }

\begin{document}
\maketitle
\begin{abstract}
Large Language Models (LLMs) have shown remarkable performance across a wide range of downstream tasks. However, it is challenging for users to discern whether the responses of LLM are generated with certainty or are fabricated to meet user expectations. In this paper, we introduce Uncertainty Tripartite Testing Paradigm (Unc-TTP), a novel method for classifying LLM uncertainty by leveraging output inconsistency. Specifically, Unc-TTP performs three rounds of sampling under varying label injection interference, enumerating all possible outcomes, and uses the degree of output inconsistency as the indicator of the LLM's intrinsic uncertainty.
To validate the effectiveness of this inconsistency-defined uncertainty, we draw inspiration from Active Learning, comparing the informativeness of actively selected in-context examples. Our experiments show that uncertainty examples selected via Unc-TTP are more informative than certainty examples. Furthermore, the Unc-TTP-guided uncertainty-based active example selection strategy outperforms existing methods, highlighting its effectiveness in classifying LLM uncertainty and enhancing in-context learning.
This work not only underscores the potential of inconsistency-based uncertainty classification for both open- and closed-source LLMs but also presents a practical approach for leveraging uncertainty to improve LLM performance in real-world tasks.

\end{abstract}

\section{Introduction}

\textit{"Real knowledge is to know the extent of one's ignorance." —— Confucius}

In the rapidly evolving field of artificial intelligence (AI), developing generic AI assistants that provide truthful and reliable information remains a challenge. 
Although Large Language Models (LLMs) are capable of fulfilling user requests for various downstream tasks, they struggle with admitting their knowledge boundary~\citep{xu2024rejection}. Additionally, they sometimes fabricate statements that are difficult to discern from falsehoods, causing trustworthiness issues~\citep{maynez2020faithfulness, alkaissi2023artificial, Survey_Hallucination}, especially when confronted with users' skepticism or misinformation~\citep{sharma2023understanding}.
These challenges push us to consider three critical questions: 
(1) \textit{How can we assess the uncertainty of an LLM’s response through its external behavior?}
(2) \textit{How can we validate the effectiveness of the assessed uncertainty?}
(3) \textit{Can we utilize the assessed uncertainty to enhance the performance of LLMs?}

Significant efforts have been devoted to investigating the external input-output behavior of LLM to assess their underlying reasoning abilities and conjecture whether LLMs have truly grasped the essence of knowledge~\citep{wang-etal-2023-chatgpt-defend, turpin_language_2023, xie2024ask, wei2024simple}. 
These studies indicate that LLMs 
display some tendencies of sycophancy, often wavering between answers when confronted with human interference. 
While current works are still trying to find ways to eliminate this sycophantic behavior, our approach takes a different direction. We capitalize on this sycophancy to develop a method for classifying uncertainty, 
leveraging the identified uncertainty to enhance LLM performance. 

We begin with a key observation from a simple experiment in which we sample LLM responses three times for the exact same instance, 
under slightly different conditions—specifically, injecting \textit{no (w/o)}, \textit{right}, or \textit{wrong} label into the prompt (Section \ref{sec:motivation}). Surprisingly, we found that although LLMs exhibit a clear tendency 
of sycophancy when faced with label injection, they still show varying degrees of resistance to this behavior. In some cases, 
LLMs consistently maintain their stance across all three samplings, with stronger models holding their answers more firmly.  Building on this 
observation, we hypothesize that \textbf{the LLM output inconsistency can serve as an indicator of its intrinsic uncertainty}.

Accordingly, we propose a fine-grained Uncertainty Tripartite Testing Paradigm (Unc-TTP), which enumerates all possible outcomes of the LLM under triple-testing scenarios \{\textit{no-label}, \textit{right-label}, \textit{wrong-label}\} and classifies LLM uncertainty levels based on the response consistency. 
Consequently, the Unc-TTP triplet assigns each instance to one of $2^3 = 8$ categories, as the LLM can either provide a correct or incorrect answer in each of the three settings. Given these eight categories, we classify any instance, 
where the LLM answers 
correct or incorrect waveringly, to be uncertain.

Basically, our proposed Unc-TTP  
is a kind of inconsistency-defined uncertainty.
Drawing inspiration from the uncertainty sampling strategy in Active Learning (AL) ~\citep{lewis1995sequential, LEWIS1994148, settles_active_2009}, 
we apply Unc-TTP’s classified uncertainty for active in-context example selection in ICL experiments. 
The preliminary study demonstrates that inconsistency-defined uncertainty examples are more informative than certainty-based ones, providing strong evidence for its 
potential for boosting LLM performance. 

Furthermore, based on three subjective text classification tasks, we conduct experiments by applying other existing strategies for active in-context example selection. Our Unc-TTP-selected uncertainty examples significantly outperform the previous best strategy, \textit{Similarity}~\cite{margatina2023activelearningprinciplesincontext}, with average accuracy improvements of 
3.7\%, 1.2\%, and 1.9\% for Llama-2, Mistral, and GPT-3.5, respectively, showcasing the effectiveness of our approach in enhancing ICL performance. We also explore the scalability and transferability of Unc-TTP classified uncertainty in selecting in-context examples, 
demonstrating its robustness and providing compelling evidence that uncertainty defined by Unc-TTP, which is calculated based on the LLM output inconsistency, 
really reveals the intrinsic uncertainty of LLMs. 


In summary, the contributions of this paper are:

\begin{itemize}
\item To the best of our knowledge, we are the first to experimentally demonstrate that output inconsistency can act as an indicator of the LLMs' intrinsic uncertainty via the perspective of active learning.

\item We propose a novel paradigm, Uncertainty Tripartite
Testing Paradigm (Unc-TTP), to test the LLM 
uncertainty for both open- and closed-source LLMs, offering a new 
insight on classifying uncertainty. 


\item 
Taking advantage of Unc-TTP, we propose an uncertainty-based active in-context example selection strategy, which yields greater performance gains than previous 
approaches in ICL. 

\end{itemize}

\section{Related Work}
\subsection{Uncertainty Behavior of LLM}
While LLMs have performed well across a variety of downstream tasks, their reliability remains a topic of concern, especially in highly specialized domains such as medicine~\citep{alkaissi2023artificial, shen2023chatgpt}. This is particularly evident when LLMs are questioned by the user and they frequently wrongly admit mistakes and give biased feedback~\citep{sharma2023understanding}. The phenomenon where a model seeks human approval in undesirable or inappropriate ways is referred to as \textit{sycophancy}~\citep{perez2022discoveringlanguagemodelbehaviors}.

Recent studies have highlighted the prevalence of sycophantic behavior in LLMs, with evidence found in various scenarios such as debates~\citep{wang-etal-2023-chatgpt-defend}, prompt manipulation to introduce misleading elements~\citep{turpin_language_2023,Unfaithful_CoT, xie2024ask, wei2024simple}, and preemptively injecting answers before engaging in Chain-of-Thought (CoT) reasoning~\citep{xu2024preemptive}. These findings collectively suggest that sycophancy may be a prevalent trait among LLMs, indicating their inability to disregard user preference and maintain a consistent stance.

While others attempt to mitigate the sycophancy of LLM \citep{sharma2023understanding, yang2023alignment, wei2024simple, wei2024measuring}, our research proposes that the wavering behavior behind sycophancy can reflect the intrinsic uncertainty of LLMs. This uncertainty, in turn, could serve as a useful indicator for performing active in-context example selection.

\subsection{ICL and Example Selection Strategy}
ICL is a paradigm that allows LLMs to learn tasks given only a few examples as demonstrations without updating model parameters~\citep{Few-Shot_Learners,dong2023survey}. In ICL, the quality of the demonstration examples is the most critical factor affecting LLM performance \citep{kumar2021reordering,lu2022fantastically}.

Research on selecting optimal in-context examples for LLMs has been an ongoing area of study. \citet{margatina2023activelearningprinciplesincontext} draws inspiration from the most common AL strategies, including uncertainty sampling~\citep{gonen2024demystifyingpromptslanguagemodels}, diversity sampling~\citep{yu2023generateretrievelargelanguage}, and similarity-based selection~\citep{liu2021makes}, to identify the most informative examples. However, these approaches fail to account for the unique characteristics of LLMs, particularly the intrinsic uncertainty revealed through sycophantic behavior. As we demonstrate in Section \ref{sec:unc-vs-cer}, this wavering behavior provides a more effective indicator for selecting in-context examples compared to traditional AL methods.

Other works have explored the use of incorrect examples, aiming to introduce more diverse information into the demonstrations. For instance, \citet{mo2024ciclcontrastiveincontextlearning} leverages incorrect examples selected through self-consistency~\citep{wang2023selfconsistency} to enhance diversity in examples. \citet{Post_Hoc_Explanations} selects incorrect examples using a weak proxy model and generates post-hoc explanations for these difficult cases. However, these methods either overlook the value of inconsistency—an important aspect that can help identify LLM blind spots—or depend on proxy models with white-box access.

In this paper, we propose a self-guided, uncertainty-based in-context example selection method that outperforms existing active ICL baselines.

\section{Preliminary and Methodology}

Despite the sycophantic tendencies of LLMs when confronted with users' skepticism or misinformation, can they disregard users' potentially bias information and maintain their stance? 
To investigate this question, we start with a simple test, whose observations lead to the development of a new methodology of classifying LLM uncertainty to improve in-context example selection: the Uncertainty Tripartite Testing Paradigm (Unc-TTP).

\subsection{A Motivating Observation}
\label{sec:motivation}

\begin{figure}[!t]
  \centering
  \begin{minipage}[b]{\linewidth}
    \centering
    \includegraphics[width=1\linewidth]{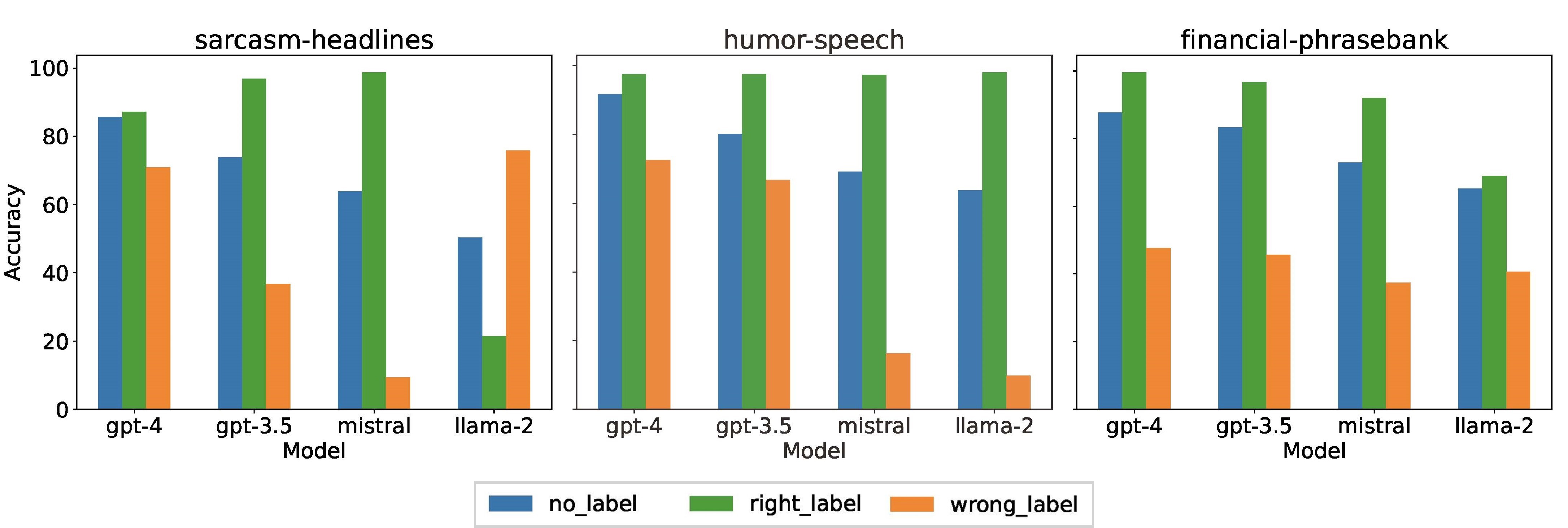}
    \caption{Accuracy of four models with and without label injection. See Appendix \ref{sec:detailbehaviour} for a detailed analysis.}
    \label{fig:bar_stage1}
  \end{minipage} \\[1ex] 
  \begin{minipage}[b]{\linewidth}
    \centering
    \begin{minipage}{0.24\linewidth}
        \centering
        \captionsetup{justification=centering}
        \subcaption*{\fontsize{6pt}{2pt}\selectfont gpt-4\\w/ label injection}
        \includegraphics[width=\linewidth,clip=true,trim=1.8cm 1.5cm 1.8cm 3cm]{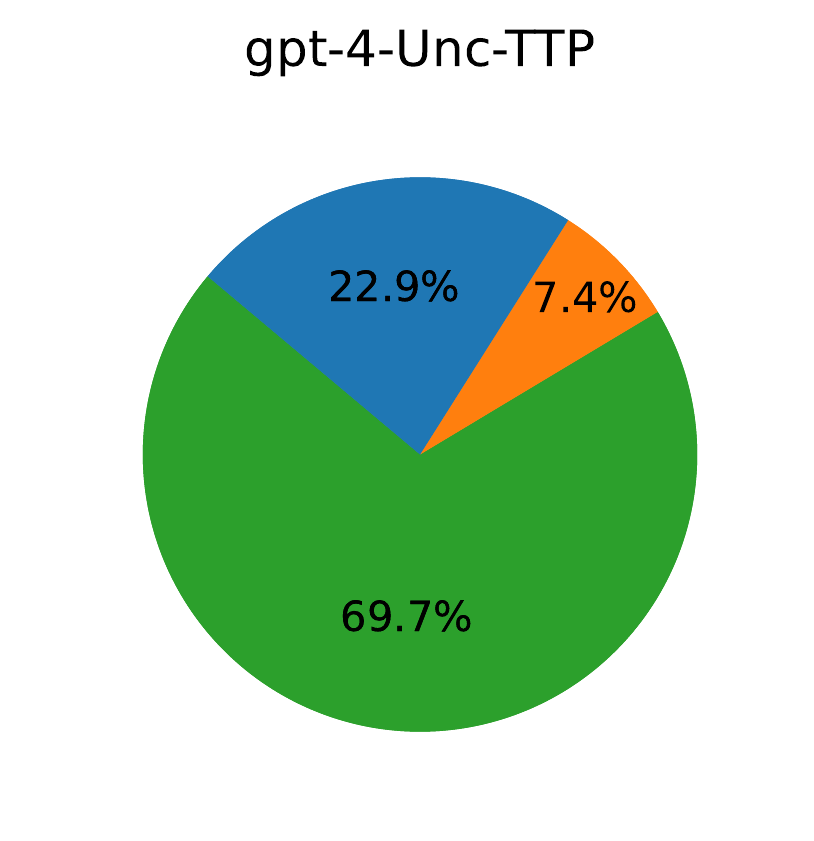}
    \end{minipage} \hfill
    \begin{minipage}{0.24\linewidth}
        \centering
        \captionsetup{justification=centering}
        \subcaption*{\fontsize{6pt}{2pt}\selectfont gpt-3.5\\w/ label injection}
        \includegraphics[width=\linewidth,clip=true,trim=1.8cm 1.5cm 1.8cm 3cm]{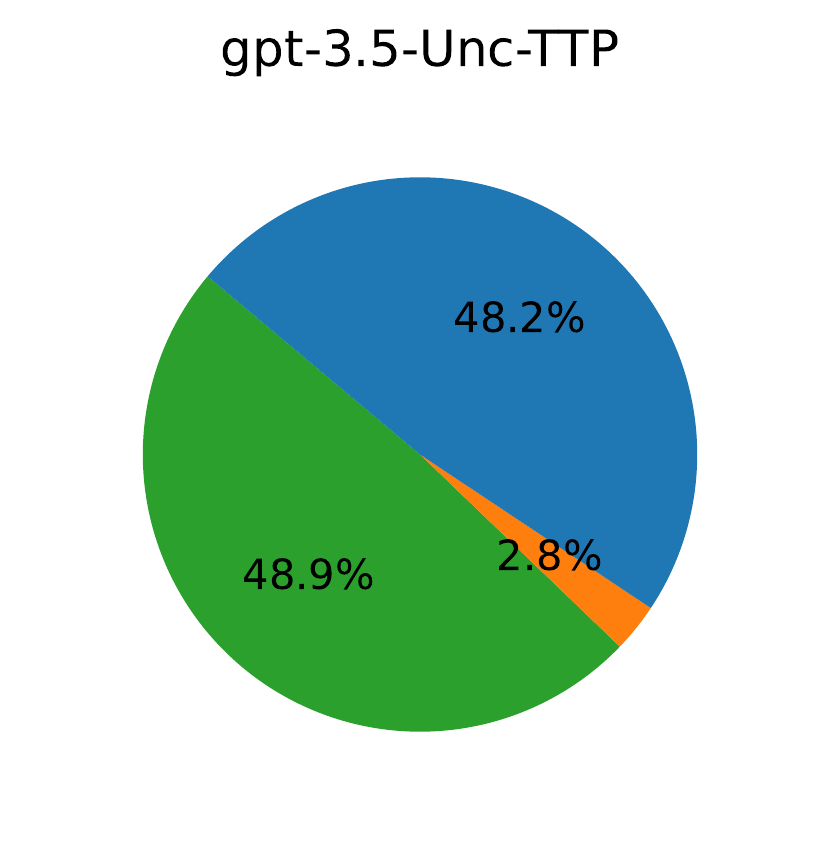}
    \end{minipage} \hfill
    \begin{minipage}{0.24\linewidth}
        \centering
        \captionsetup{justification=centering}
        \subcaption*{\fontsize{6pt}{2pt}\selectfont llama-2\\w/ label injection}
        \includegraphics[width=\linewidth,clip=true,trim=1.8cm 1.5cm 1.8cm 3cm]{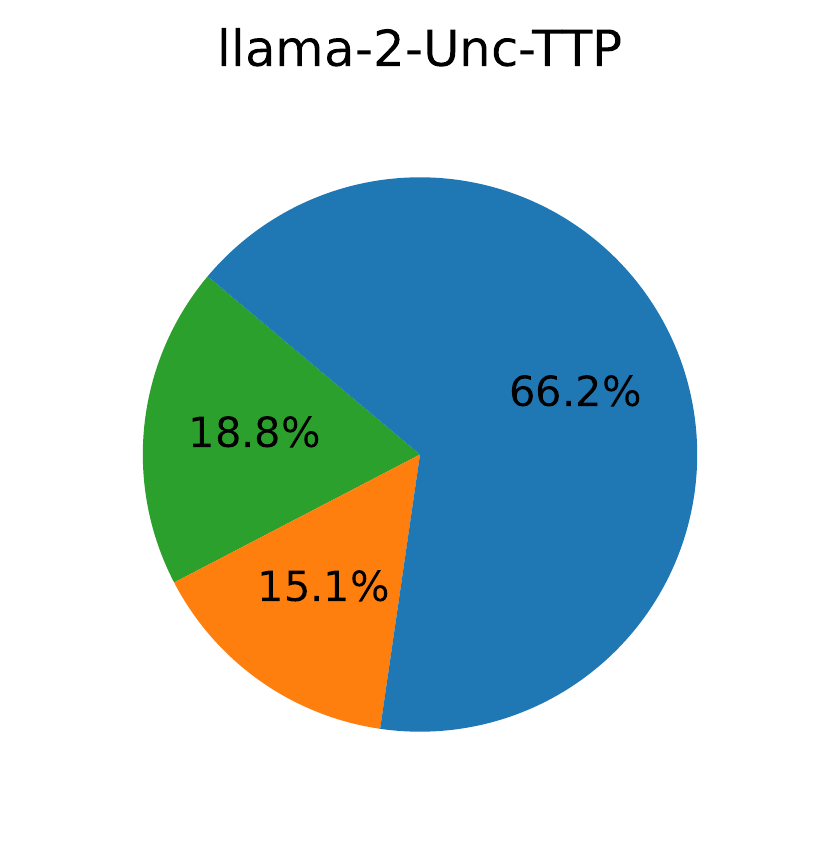}
    \end{minipage} \hfill
    \begin{minipage}{0.24\linewidth}
        \centering
        \captionsetup{justification=centering}
        \subcaption*{\fontsize{6pt}{2pt}\selectfont mistral\\w/ label injection}
        \includegraphics[width=\linewidth,clip=true,trim=1.8cm 1.5cm 1.8cm 3cm]{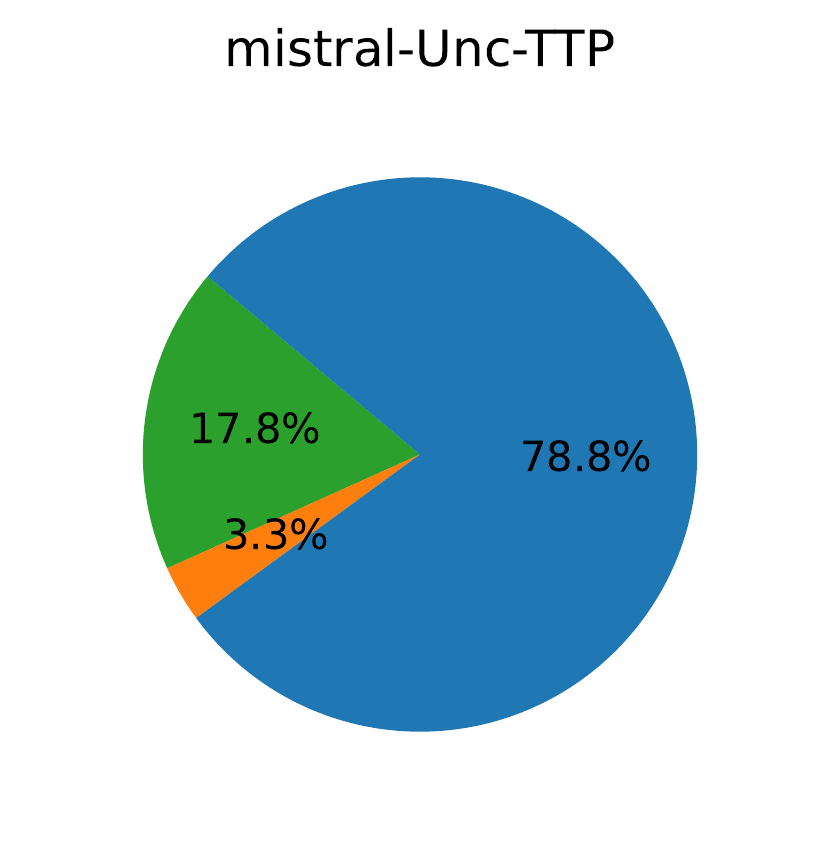}
    \end{minipage} \hfill
    
    \begin{minipage}{0.24\linewidth}
        \centering
        \subcaption*{\fontsize{6pt}{2pt}\selectfont gpt-4-vanilla}
        \includegraphics[width=\linewidth,clip=true,trim=1.8cm 1.5cm 1.8cm 3cm]{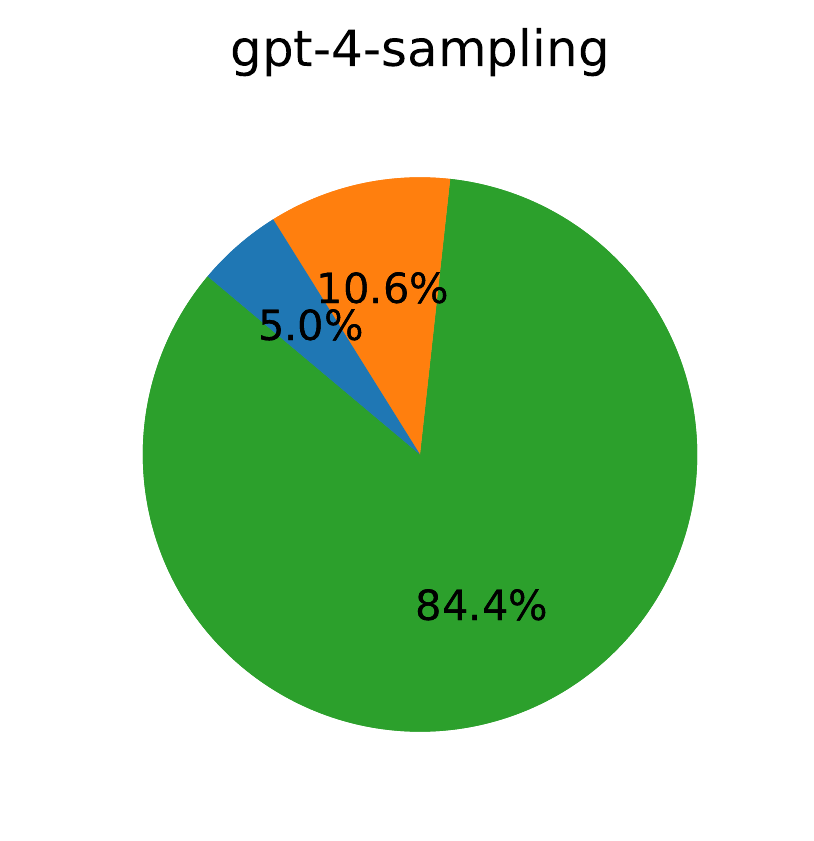}
    \end{minipage} \hfill
    \begin{minipage}{0.24\linewidth}
        \centering
        \subcaption*{\fontsize{6pt}{2pt}\selectfont gpt-3.5-vanilla}
        \includegraphics[width=\linewidth,clip=true,trim=1.8cm 1.5cm 1.8cm 3cm]{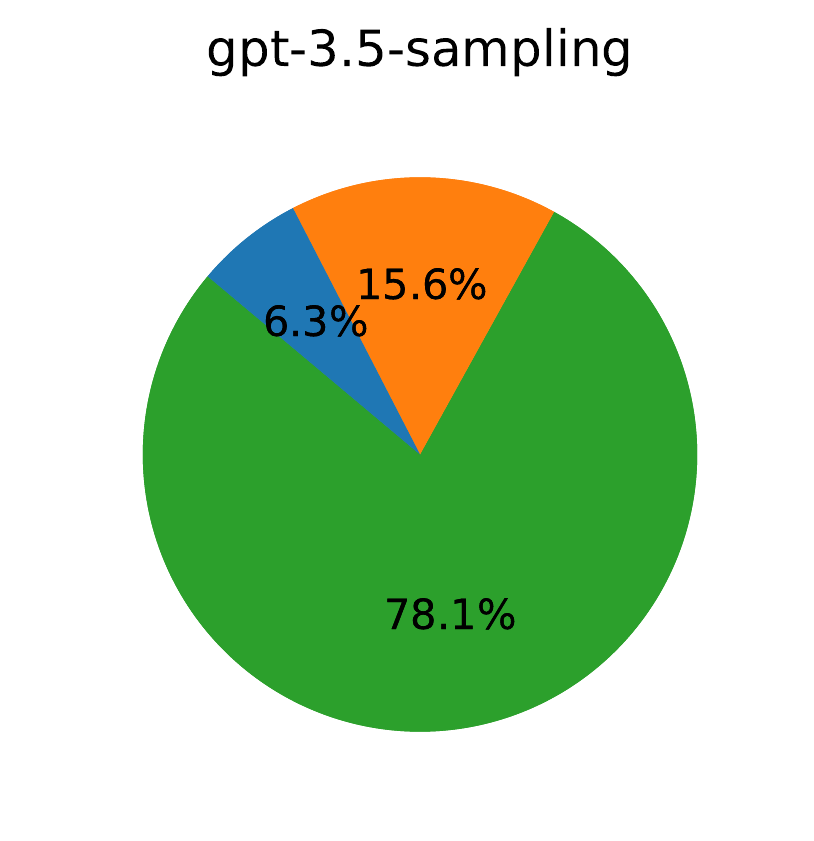}
    \end{minipage} \hfill
    \begin{minipage}{0.24\linewidth}
        \centering
        \subcaption*{\fontsize{6pt}{2pt}\selectfont llama-2-vanilla}
        \includegraphics[width=\linewidth,clip=true,trim=1.8cm 1.5cm 1.8cm 3cm]{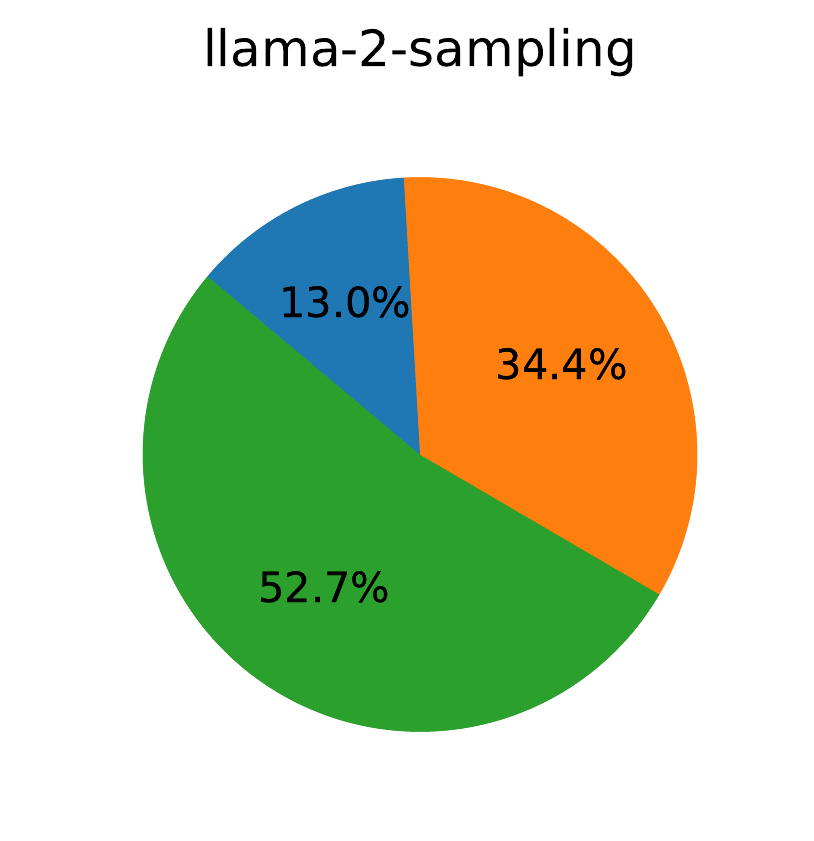}
    \end{minipage} \hfill
    \begin{minipage}{0.24\linewidth}
        \centering
        \subcaption*{\fontsize{6pt}{2pt}\selectfont mistral-vanilla}
        \includegraphics[width=\linewidth,clip=true,trim=1.8cm 1.5cm 1.8cm 3cm]{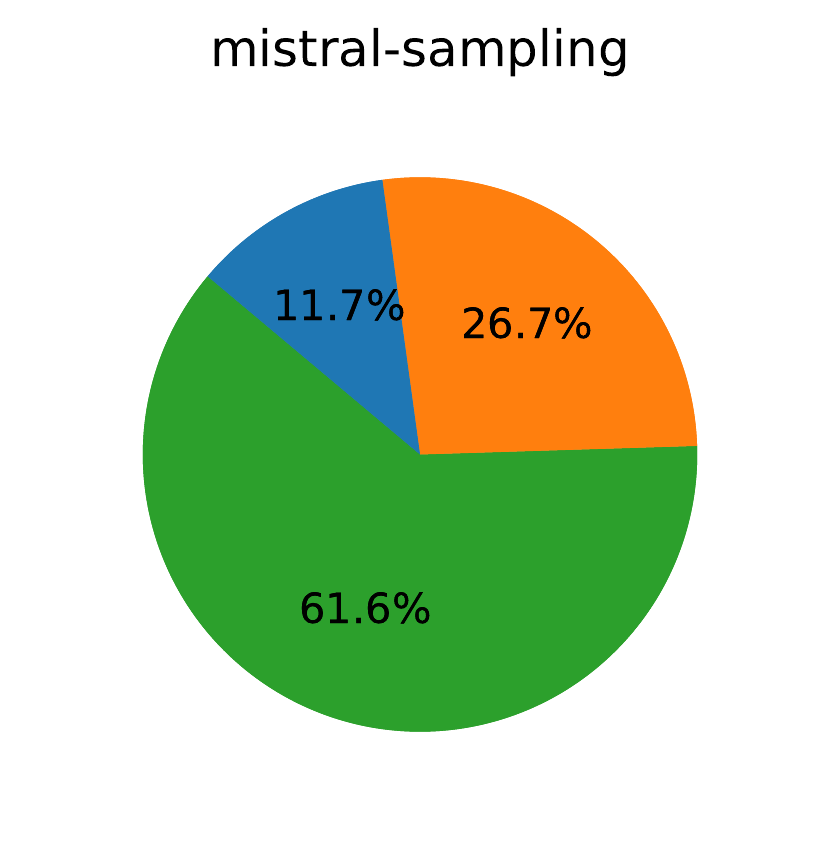}
    \end{minipage} \hfill
    \begin{minipage}{\linewidth}
        \centering
        \vspace{0.5em}
        \includegraphics[width=0.8\linewidth]{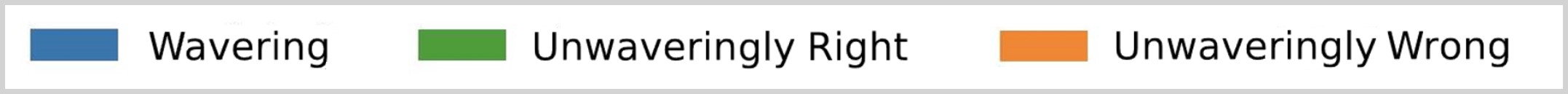}
    \end{minipage} \hfill

    \caption{Distribution of instances where the LLM remains unwavering (certain) and wavering (uncertain) across three rounds of sampling, with \{\textit{no (w/o)}, \textit{right}, \textit{wrong}\} label injection (top) and without label injection, using only vanilla temperature sampling (bottom), across three datasets.}
    \label{fig:pie_stage1}
  \end{minipage}
\end{figure}

The simple test is conducted under three settings: We prompt LLMs with regular query, query with \textit{right} label and query with \textit{wrong} label, using the prompt shown in Table \ref{tab:prompt}, the LLMs and datasets described in Sections \ref{sec:modelConfig} and \ref{sec:datasets}. 
The accuracy of four models is shown in Figure~\ref{fig:bar_stage1}. 
Surprisingly, we observed varying degrees of resistance to sycophantic behavior in LLMs during their decision-making processes. In some cases, even when provided with \textit{right} or \textit{wrong} labels interference, the LLMs maintained their stance unwaveringly, revealing that, in some particular instances, LLMs can firmly maintain their stance across all three settings. This behavior might reflect their intrinsic uncertainty and certainty.

To further investigate this potential uncertainty (define by output inconsistency) across different LLMs, we present the distribution of wavering and unwavering instances for each model, selected using the settings described above (top), alongside the results from the vanilla method, which samples outputs three times under the no label injection setting using temperature sampling (bottom), as shown in Figure~\ref{fig:pie_stage1}.

Derived from the results of Figure~\ref{fig:pie_stage1}, we make the following observations.
Stronger LLMs exhibit greater output consistency, with a larger percentage of them being able to unwaveringly maintain their answers across three samplings.
We attribute this to the LLM's confidence in its answers. Stronger models possess better logical reasoning ability, allowing them to maintain their stance regardless of label injection or temperature sampling. Moreover, compared to the vanilla sampling-based method, injecting \textit{right}/\textit{wrong} label into the prompt, making it much more difficult for LLMs to persist in its answer, therefore improves the strictness of identifying the actual instances where LLMs are certain.

Building on these key observations, we hypothesize that \textbf{the LLM output inconsistency can serve as an indicator of its uncertainty}. In order to prove the validity of this hypothesis, we draw inspiration from uncertainty sampling in AL and adopt AL principle for in-context example selection, where we select the most informative instances with our \textbf{inconsistency-defined uncertainty} as selection strategy. Since LLMs exhibit varying behaviors for different injected labels (see Appendix \ref{sec:detailbehaviour}), this underscores the need for a fine-grained uncertainty classification method. To address this, we propose Unc-TTP.

\begin{table}[!htbp]
    \centering
    \begin{tabular}{p{7.2cm}}
    \toprule
        \textbf{Prompt} \\
        \midrule
        \textbf{\textit{no} label setting}  (w/o label injection)
        
        Your job is to determine whether the text is \{A\} or \{B\}. \\
        \midrule
        \textbf{\textit{right} / \textit{wrong} label setting} \\
         Your job is to determine whether the text is \{A\} or \{B\} by reference to the given label, which presents the ground truth of the text as \{A\} or \{B\}. Despite having to refer to the provided labels, you should still have your own thinking and do not change your stance so easily. \\
        \bottomrule
    \end{tabular}
    \caption{Prompts used in Unc-TTP, where \{A\} and \{B\} denote the actual label from the training data.}
\label{tab:prompt}
\end{table}


\subsection{The Uncertainty Tripartite Testing Paradigm}

\begin{figure*}[!t]
  \centering
  \includegraphics[width=\textwidth]{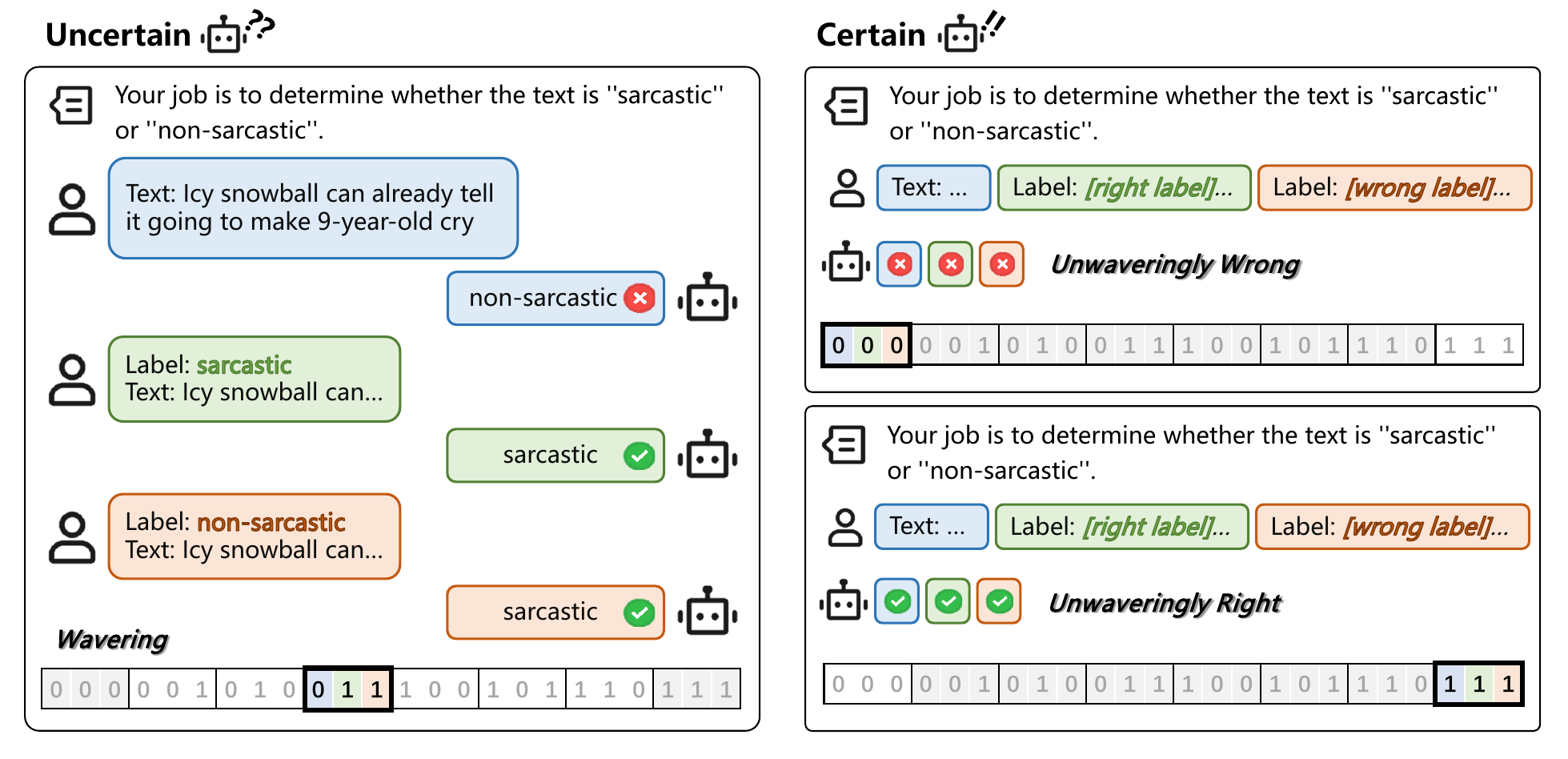}
  \caption {Illustration of the proposed Uncertainty Tripartite Testing Paradigm (Unc-TTP) on the Sarcasm Headlines (SH) dataset. For each instance, we employ Unc-TTP to evaluate the LLM's certainty level and label it as either \textit{certain} or \textit{uncertain} based on its combination of testing scenarios. Each category consists of three components, corresponding to the three individual testing scenarios in the order of \{\textit{no-label}, \textit{right-label}, \textit{wrong-label}\}. If the LLM answers incorrectly under a given setting, it is labeled as \textit{0}; otherwise, it is labeled as \textit{1}. We interpret instances where the LLM wavers between answers as indicative of \textit{uncertain}. Conversely, instances where the LLM is unwaveringly right or wrong are considered \textit{certain}.}
  \label{fig:pipeline}
\end{figure*}

The schematic of our proposed Unc-TTP is illustrated in Figure~\ref{fig:pipeline}. It involves asking LLMs to answer a question under three independent settings: (1) providing with \textit{no} (w/o) label, (2) providing with the \textit{right} label, and (3) providing with the \textit{wrong} label, where the labels are sourced from benchmark datasets. We denote this tripartite paradigm as \textit{\{no-label, right-label, wrong-label\}} for the sake of brevity. 

At the same time, to encourage the model to think independently, we prompt the model to keep its own stances when facing label interference, and the specific prompt is shown in Table \ref{tab:prompt}. 

The rationale behind \textit{no-label} setting is that we want to get the model answer that is derived entirely from the LLM's own knowledge and reasoning ability, without any external guiding interference. As for the \textit{right-label} and \textit{wrong-label} settings, we provide LLM with either a right or wrong label, intending to lure the model into making the decision we are steering it towards. These settings expose areas where the model is not certain.


\paragraph{Uncertainty Category Labeling}
\label{category_labeling}
After testing all three settings, for each instance, we obtain three results corresponding to \textit{\{no-label, right-label, wrong-label\}}, each with two possible outcomes: the model answering correctly or incorrectly. Marking the instances that the model answers correctly as \textit{1} and incorrectly as \textit{0}, we get a total of eight possible outcomes, denoted as \textit{\{000, 001, 010, 011, 100, 101, 110, 111\}}. These eight categories can be further grouped into two types:

\noindent \textbf{Certain:} When the model provides consistent answers for an instance under all three settings, we consider the model to be certain with respect to the instance. The categories \textit{000} (denoted as Cer$_{W}$), and \textit{111} (denoted as Cer$_{R}$) fall into this type, where the model’s outputs are \textit{unwaveringly wrong} in the former and \textit{unwaveringly right} in the latter.

\noindent \textbf{Uncertain:}
In the cases where the model provides \textit{wavering} answers, including \textit{001}, \textit{110}, \textit{011}, \textit{100}, \textit{101}, and \textit{110}, we consider the model to be uncertain about the instance. Taking \textit{011} as an example, it represents a data instance where the model answers incorrectly with the \textit{no-label} setting and correctly with the \textit{right-label} and \textit{wrong-label} settings. We denote these instances as Uncertain (Unc).

\subsection{Example Selection with Unc-TTP}
Taking advantage of Unc-TTP, we assign each instance in the training set with a specific uncertainty categories, and the category distribution is shown in Figure~\ref{fig:heat-map}. The distribution reveals significant unevenness, indicating that each LLM and dataset possesses a unique uncertainty property.
In this section, we select examples based on the fine-grained uncertainty categories classified by Unc-TTP.

We select the instances independently from each of the categories classified by Unc-TTP
as the in-context examples to perform $K$-way $N$-shot ICL, where $K$ denotes the number of the labels, and $N$ denotes the sample numbers.
Due to the uneven distribution of sample quantities across different categories, some categories might have insufficient samples to complete the $K$-way $N$-shot ICL. In such cases, we will supplement with randomly selected examples. If one category has no corresponding instance at all, we will drop it. Therefore, we choose 1-shot as our main experiment since a smaller $N$ can better reflect the true impact of uncertainty categories. 


\section{Experimamtal Setup}

\begin{figure*}[!t]
    \centering
    \setlength{\tabcolsep}{0.8mm}
      \includegraphics[height=3cm,width=0.25\linewidth]{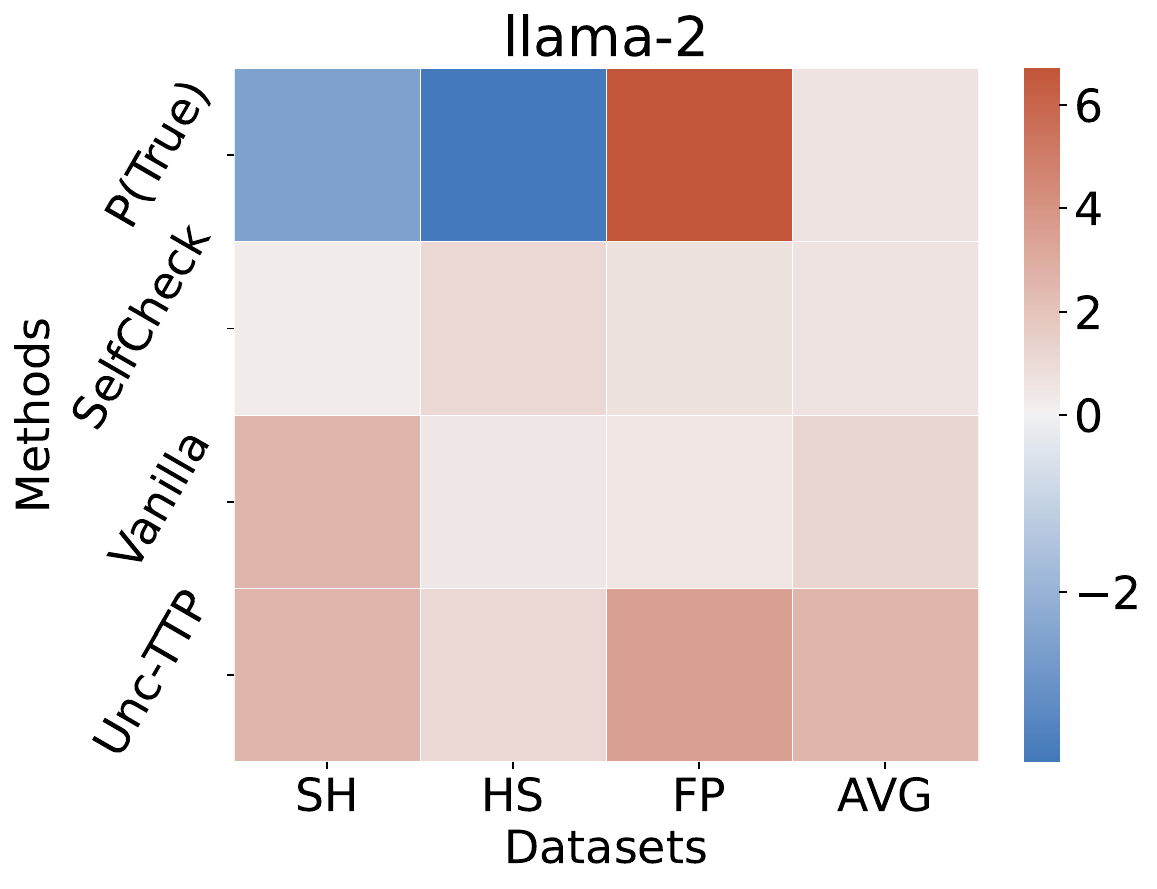}\hspace{0em}
      \includegraphics[height=3cm,width=0.25\linewidth,clip=true,trim=1cm 0cm 0cm 0cm]{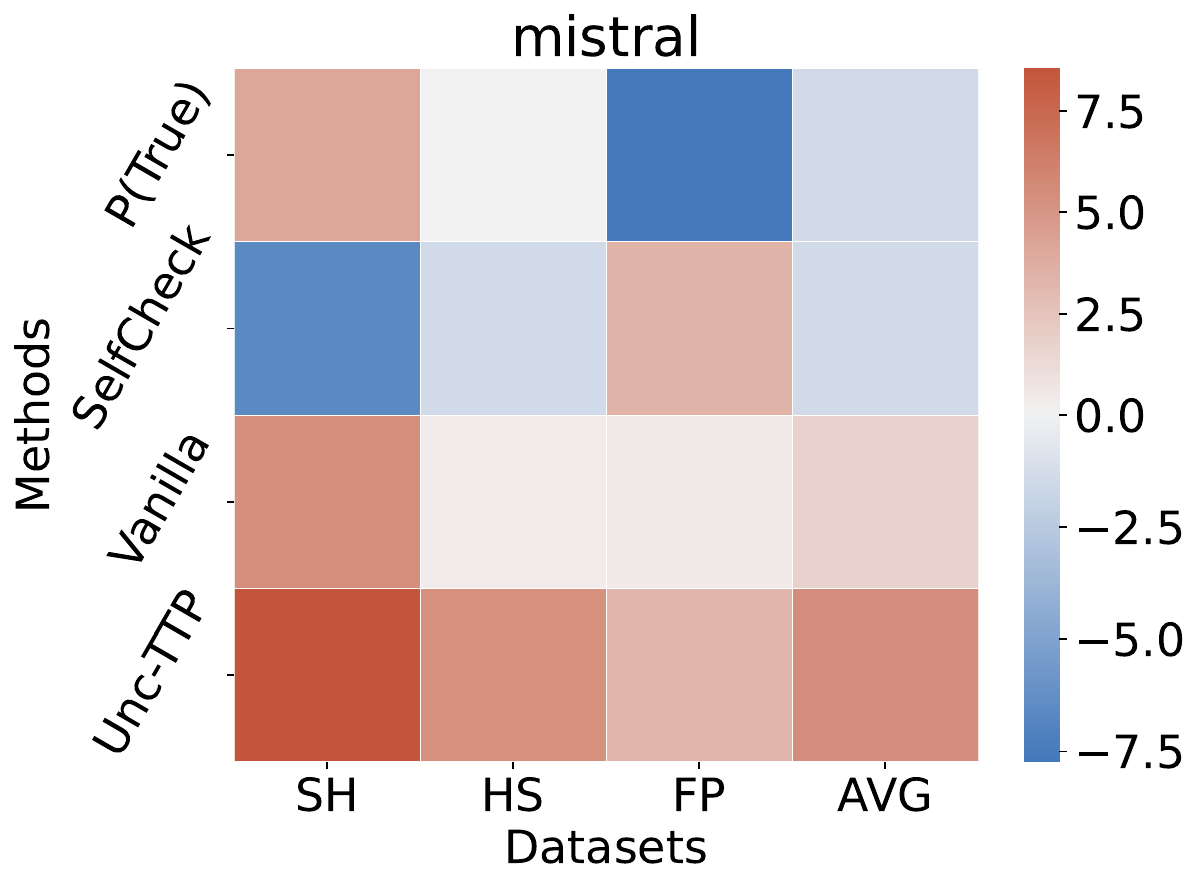}\hspace{0em}
      \includegraphics[height=3cm,width=0.25\linewidth,clip=true,trim=1cm 0cm 0cm 0cm]{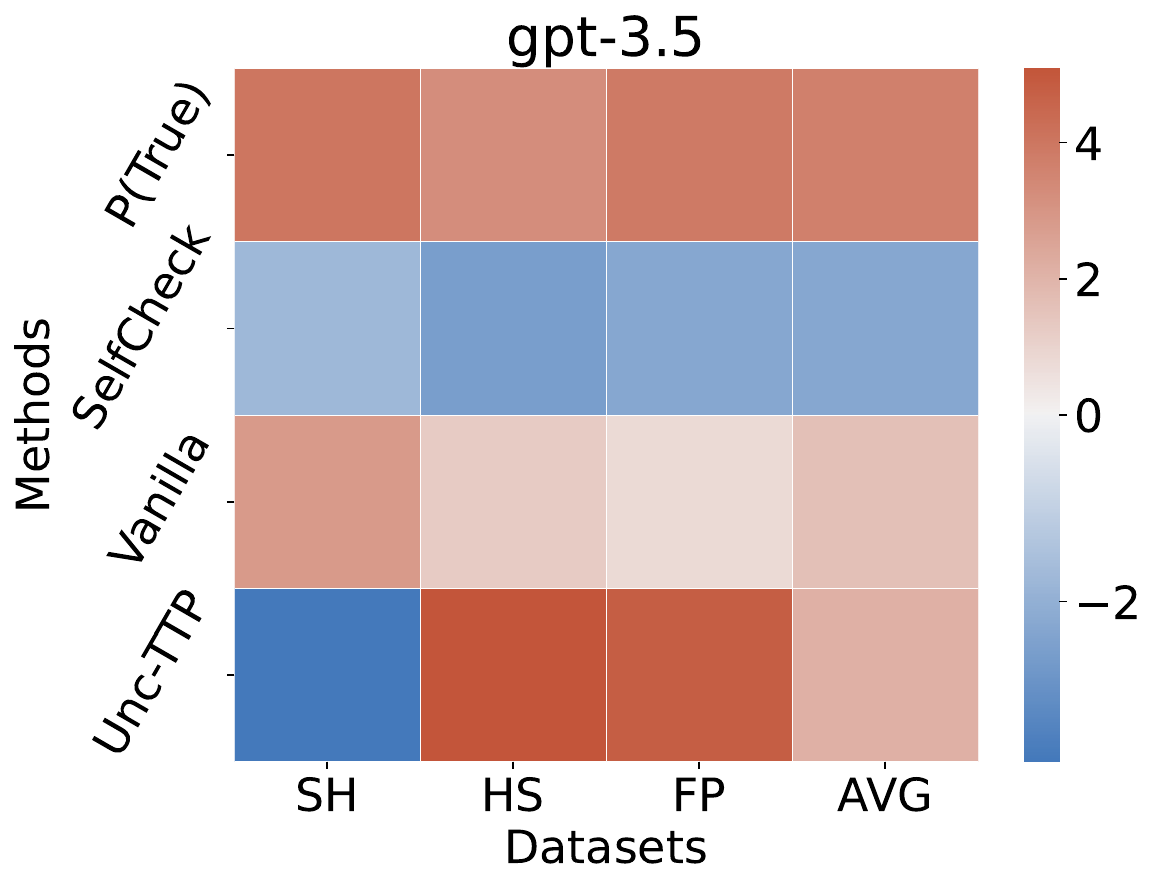}\hspace{0em}
      \caption {The relative improvement in accuracy of inconsistency-defined uncertainty examples compared to certainty examples in ICL for each model. The uncertainty examples from sampling-based methods (the bottom 2 rows) demonstrate greater robustness than those from verification-based methods (the top 2 rows), where positive accuracy gains (highlighted in red) are observed more consistently than negative ones (highlighted in blue).}
    \label{fig:relative_gain}
\end{figure*}

\subsection{Model Configurations}
\label{sec:modelConfig}
We perform our experiments with GPT-3.5 and GPT-4
\footnote{Our experiments were done with \textit{gpt-3.5-turbo-1106} and the \textit{gpt-4-0613} checkpoint. The results for certain examples may change due to model updates.}
~\citep{openai2024gpt4}, llama-2-7b-chat~\citep{touvron2023llama} and mistral-7b-instruct~\citep{jiang2023mistral}, a model that significantly outperforms the llama-2-13b-chat ~\citep{jiang2023mistral}. 
All generations are done under the temperature of 0.7 except Unc-TTP example selection is done via greedy decoding.

\subsection{Datasets}
\label{sec:datasets}
We select three representative subjective text classification tasks for our experiments, according to the subjectivity ranking~\cite{li_synthetic_2023}. 
\textbf{Sarcasm Headlines} (SH) \citep{misra_sarcasm_2023}, a high-quality news headlines sarcasm detection dataset.
\textbf{Humor Speech} (HS) \citep{Annamoradnejad_2024}, a dataset consisting of short texts with both humorous and non-humorous content. 
\textbf{Financial Phrasebank} (FP) \citep{malo_good_2014}, a sentiment classification task that divides finance-related sentences as positive, negative, or neutral. 
Please refer to Appendix~\ref{sec:dataset-detail} for a detailed data description.

\subsection{Comparing Example Selection Methods}
\subsubsection{Output Inconsistency-based Methods}
To evaluate the validity of inconsistency-defined uncertainty and compare the informativeness of uncertainty versus certainty instances, we implement two types of inconsistency measurement methods: Verification-based and Sampling-based methods, with $q=3$ to align with the Unc-TTP tripartite testing paradigm.

\paragraph{Verification-based Methods}
These two-stage methods measure consistency between the LLM's answer and its self-verification, checking if the model consistently supports its initial response. We adopt \textbf{SelfCheckGPT} \cite{manakul-etal-2023-selfcheckgpt} and \textbf{P(True)} \cite{kadavath2022languagemodelsmostlyknow}, both of which are usually used in fact-checking tasks. \textbf{SelfCheckGPT} scores each instance based on the consistency across $q$ rounds of verification. The higher scores indicate greater uncertainty (denoted as Unc$_S$), while lower scores reflect higher certainty (denoted as Cer$_S$). \textbf{P(True)} calculates the probability that the LLM outputs "\textit{True}" during the verification stage, where lower probabilities indicate uncertainty, and higher probabilities suggest certainty.

\paragraph{Sampling-based Methods}
These assess output inconsistency through multiple sampling iterations, with Unc-TTP falling into this category. The \textbf{Vanilla Sampling} (Vanilla) method, also known as temperature sampling, inspired by self-consistency \cite{wang2023selfconsistency}, uses $q$ decoding paths followed by majority voting to identify instances of highest certainty. We extend this approach by classifying output from the $q$ samplings into the categories of Cer$_R$, Cer$_W$, and Unc.

\subsubsection{Traditional Methods}
To evaluate the effect of our proposed Unc-TTP-guided uncertainty
as an active in-context example selection strategy, we compare our method with the \textbf{Random} baseline as well as other common AL approaches, following \citet{margatina2023activelearningprinciplesincontext}.

\textbf{Similarity} \cite{liu2021makes} 
selects the top-$k$ semantically similar examples 
as demonstrations for each test sample.
\textbf{BM25} \cite{INR-019} ranks training examples based on their relevance to each test instance, and the top-ranked examples are selected as ICL demonstrations.
\textbf{Diversity} \cite{margatina2023activelearningprinciplesincontext} encodes all train data with Sentence-BERT \cite{2019sentencebert} and performs $k$-means clustering. We select one instance from each cluster. 
\textbf{Perplexity} and \textbf{Entropy} are two classical measures to evaluate LLMs' uncertainty properties \cite{margatina2023activelearningprinciplesincontext}.

\subsection{Implementation Details}
We conduct the example selection exclusively on the training set, ensuring that we remain blind to both the validation and test sets at this stage. After that, we evaluate these candidates on the validation set to determine which uncertainty category performs best and adopt the best-performing category example for ICL testing.
All ICL experiments are repeated three times with different random seeds, and we report the average accuracy along with the standard deviation. As for the prompt for evaluating the baseline method, we use the same as the one with \textit{no-label} setting.




\section{Experimental Results and Analysis}

First, we conduct experiments with several methods that measure output inconsistency to demonstrate that 
uncertain instances are more informative than the certain ones when used as in-context examples. Subsequently, we compare our 
Unc-TTP-guided uncertainty sampling strategy with other common baselines, to further illustrate the validity of our method.

\subsection{Preliminary Study: Uncertain Versus Certain Instances}
\label{sec:unc-vs-cer}

\begin{table}[!t]
    \centering
    \small
    \begin{tabular}{c|lccc}
    \toprule
        \textbf{Methods} & \multicolumn{1}{c}{\textbf{Cat.}} & \textbf{Llama-2} & \textbf{Mistral} & \textbf{GPT-3.5} \\
        \cmidrule{1-5}
        \multicolumn{5}{l}{\textbf{Verification-Based}} \\ [3pt] 
        \multirow{2}{*}{P(True)} & Cer$_P$ & 66.6 (1.1) & \underline{77.5} (1.2) & 81.4 (1.2) \\
        ~ & Unc$_P$ & \underline{67.0} (1.0) & 76.4 (0.8) & \underline{84.4} (1.3) \\
    \cmidrule{1-5}
        SelfCheck & Cer$_{S}$ & 66.7 (1.7) & \underline{77.5} (2.4) & \underline{83.8} (1.3) \\
        GPT & Unc$_S$ & \underline{67.1} (1.6) & 76.4 (2.2) & 81.9 (1.6) \\ 
    \cmidrule{1-5} 
    \multicolumn{5}{l}{\textbf{Sampling-Based}} \\ [3pt] 
        \multirow{2}{*}{Vanilla} & Cer$_W$ & 66.7 (1.7) & 78.1 (2.9) & 84.4 (1.9) \\
        ~ & Cer$_R$ & 66.9 (2.0) & 78.2 (3.4) & 83.7 (2.0) \\
        Sampling & Unc & \underline{67.6} (1.1) & \underline{79.5} (1.2) & \underline{\textbf{85.4}} (2.9) \\
    \cmidrule{1-5}
        \multirow{3}{*}{Unc-TTP} & Cer$_W$ & 69.0 (4.4) & 79.5 (3.4) & 82.3 (2.3) \\
        ~ & Cer$_R$ & 68.1 (2.7) & 75.2 (2.2) & 84.6 (2.4) \\
        ~ & Unc & \underline{\textbf{70.3}} (2.2) & \underline{\textbf{81.5}} (1.5) & \underline{85.2} (1.5) \\
    \bottomrule
    \end{tabular}
    \caption{The average accuracy on three datasets for each model over four methods that measure uncertainty via output
    inconsistency.} 
    \label{tab:main-unc-cer-test-avg}
\end{table}

\begin{table*}[!tbp]
    \centering
    \setlength{\tabcolsep}{1.8mm} 
    \tiny
    \begin{tabular}{lcccc|cccc|cccc}
    \toprule
        \textbf{Methods} & \multicolumn{4}{c}{\textbf{Llama-2}} & \multicolumn{4}{c}{\textbf{Mistral}} & \multicolumn{4}{c}{\textbf{GPT-3.5}} \\
        \cmidrule(r){2-5} \cmidrule(r){6-9} \cmidrule(r){10-13} 
        ~ & \textbf{SH} & \textbf{HS} & \textbf{FP} & \textbf{Avg.} & \textbf{SH} & \textbf{HS} & \textbf{FP} & \textbf{Avg.} & \textbf{SH} & \textbf{HS} & \textbf{FP} & \textbf{Avg.} \\
        \midrule
        \multicolumn{12}{l}{
        \textbf{Random Baseline}} \\  [3pt]
        Random & 69.8 (1.6) & 53.5 (1.5) & 78.3 (1.8) & 67.2 (1.6) & 71.2 (1.6) & \underline{81.6} (4.4) & 84.6 (2.3) & 79.1 (2.8) & \textbf{\underline{83.1}} (4.3) & \underline{88.8} (1.8) & 83.2 (0.8) & \underline{85.0} (2.3) \\ 
        \midrule
        \multicolumn{12}{l}{
        \textbf{Traditional Active Learning Methods}} \\  [3pt]
        Similarity* & \underline{69.9} (0.7) & 52.7 (1.5) & 77.3 (0.0) & 66.6 (0.7) & \textbf{\underline{76.0}} (0.5) & 78.8 (0.6) & \underline{86.1} (0.3) & \underline{80.3} (0.5) & \underline{79.2} (0.6) & \underline{87.3} (0.6) & \underline{83.4} (0.2) & \underline{83.3} (0.5) \\ [1pt]
        BM25* & 68.0 (0.7) & 55.4 (1.1) & 74.7 (1.2) & 66.0 (1.0) & 72.8 (0.6) & 80.0 (1.1) & 85.5 (0.9) & 79.4 (0.9) & 74.8 (1.6) & 87.2 (1.0) & 83.2 (1.0) & 81.7 (1.2) \\ [1pt]
        Diversity & 66.5 (1.0) & \underline{55.9} (0.7) & \textbf{\underline{81.5}} (0.6) & \underline{68.0} (0.8) & 53.7 (0.9) & 81.1 (1.9) & 84.1 (0.6) & 73.0 (1.1) & 62.7 (1.6) & 86.7 (0.9) & 81.7 (1.1) & 77.0 (1.2) \\ [1pt]
        Entropy\ding{61} & 68.5 (0.8) & 52.7 (0.1) & 80.2 (0.9) & 67.1 (0.6) & 68.0 (0.4) & \underline{81.5} (0.7) & 83.0 (0.7) & 77.5 (0.6) & - & - & - & - \\ [1pt]
        Perplexity\ding{61} & 66.3 (1.3) & 54.9 (1.6) & 80.5 (0.7) & 67.2 (1.2) & 69.0 (0.7) & 79.5 (0.3) & 85.2 (1.3) & 77.9 (0.8) & - & - & - & - \\ [1pt]
        \midrule [1pt]
        \multicolumn{12}{l}{\textbf{Inconsistency-Defined Uncertainty Based Active Learning Methods}} \\  [3pt]
        P(true) & 68.2 (0.8) & 51.5 (1.2) & \underline{81.2} (1.1) & 67.0 (1.0) & 73.2 (1.0) & 80.7 (0.6) & 75.2 (0.8) & 76.4 (0.8) & 77.8 (1.0) & 92.2 (0.6) & 83.3 (2.2) & 84.4 (1.3) \\ [1pt]
        SelfCheckGPT & 66.2 (1.7) & 57.3 (1.9) & 77.8 (1.1) & 67.1 (1.6) & 65.5 (2.9) & 80.8 (2.1) & 82.9 (1.5) & 76.4 (2.2) & 77.3 (3.5) & 87.5 (0.7) & 81.0 (0.7) & 81.9 (1.6) \\ [1pt]
    Vanilla Sampling & 69.1 (1.0) & 53.8 (1.0) & 79.8 (1.2) & 67.6 (1.1) & 72.0 (0.9) & 79.5 (1.4) & 87.0 (1.2) & 79.5 (1.2) & \underline{80.1} (5.5) & 92.5 (1.8) & 83.5 (1.3) & \textbf{\underline{85.4}} (2.9) \\ [1pt]
        \cellcolor{gray!20}Unc-TTP & \cellcolor{gray!20}\textbf{\underline{71.2}} (2.1) & \cellcolor{gray!20}\textbf{\underline{58.8}} (3.2) & \cellcolor{gray!20}80.8 (1.3) & \cellcolor{gray!20}\textbf{\underline{70.3}} (2.2) & \cellcolor{gray!20}\underline{74.7} (0.8) & \cellcolor{gray!20}\textbf{\underline{81.8}} (2.9) & \cellcolor{gray!20}\textbf{\underline{88.1}} (0.8) & \cellcolor{gray!20}\textbf{\underline{81.5}} (1.5) & \cellcolor{gray!20}77.8 (2.4) & \cellcolor{gray!20}\textbf{\underline{93.0}} (0.5) & \cellcolor{gray!20}\textbf{\underline{84.8}} (1.7) & \cellcolor{gray!20}85.2 (1.5) \\ [1pt]
    \bottomrule
    \end{tabular}
    \caption{1-shot results of LLMs on three subjective tasks in the test set. The table presents the mean accuracy across three estimations, with standard deviations in parentheses. The highest accuracy values under each method are underlined. The best results for each dataset are highlighted in bold. Please note that * indicates methods that select different examples for each test instance, while \ding{61} signifies methods that require white-box access, which is not compatible with the closed-source LLM GPT-3.5.}
    \label{tab:compare2albaseline}
\end{table*}

To justify the uncertainty defined by inconsistency, we draw on the concept of uncertainty sampling in AL and apply inconsistency-defined uncertainty on in-context example selection to evaluate its effect.
The results are shown in Table \ref{tab:main-unc-cer-test-avg} and visualized in Figure \ref{fig:relative_gain}.

Verification-based methods try to measure the consistency between the LLM's first stage answer and the second stage self-verification to reveal LLM uncertainty. In this case, risking more possibility of sycophancy and cannot reflect the intrinsic uncertainty of LLM. Therefore showing no clear pattern of the relative improvement in accuracy of uncertainty examples compared to certainty examples in ICL for each model.

However, sampling-based methods, vanilla sampling and our Unc-TTP, select uncertainty instances which bring consist performance gain compare to the certain ones in most of the cases, showcasing that the selected uncertain instances are indeed more informative than the certain one under the context of ICL. Moreover, compared to vanilla, the relative performance gain is more significant for the uncertain example classified by our proposed Unc-TTP approach, which introduces label injection to strengthen the sampling rigor. 
As for the exception, SH on GPT-3.5, we will discuss it on the following section.

\subsection{Main Results: Comparing Unc-TTP with Previous AL Methods}
\label{sec:compare2albaseline}

We have shown that the uncertainty examples selected through our Unc-TTP are more informative than the certainty examples on ICL. In this section we continue to explore the effectiveness of our proposed Unc-TTP compared to other AL strategies. The results are shown in Table \ref{tab:compare2albaseline}. 

For applying the traditional AL strategy on ICL, \textit{Similarity} have demonstrated a obvious advantage over other strategy on GPT-3.5 and Mistral, even with the more common strategy like \textit{Entropy} in traditional AL, which align with the previous results of \citet{margatina2023activelearningprinciplesincontext}. It suggests that the more capable models rely more on sample similarity while weaker model like Llama-2 depends more on example \textit{Diversity}. Yet, we interestingly discover, compare to other traditional AL strategies, \textit{Random} are the most informative on the strongest model GPT-3.5. It indicates that strong LLM may be insensitive to examples.

For inconsistency-defined uncertainty as the active in-context example selection strategy, ours Unc-TTP demonstrate a competitive advantage. Moreover, Unc-TTP not only performs better than the AL baseline \textit{Random} by the average of 3.1\%, 2.4\% and 0.2\% on Llama-2, Mistral and GPT-3.5, but also far exceeds the best strategy \textit{Similarity} stated in previous work \citet{margatina2023activelearningprinciplesincontext} with the improvement of 3.7\%, 1.2\% and 1.9\%, demonstrating the effectiveness of our Unc-TTP on selecting the most informative in-context examples by the inconsistency-defined uncertainty.

\subsection{Assessing the Scalability of Unc-TTP}

\begin{figure*}[!t]
\centering
  \includegraphics[width=0.25\linewidth]{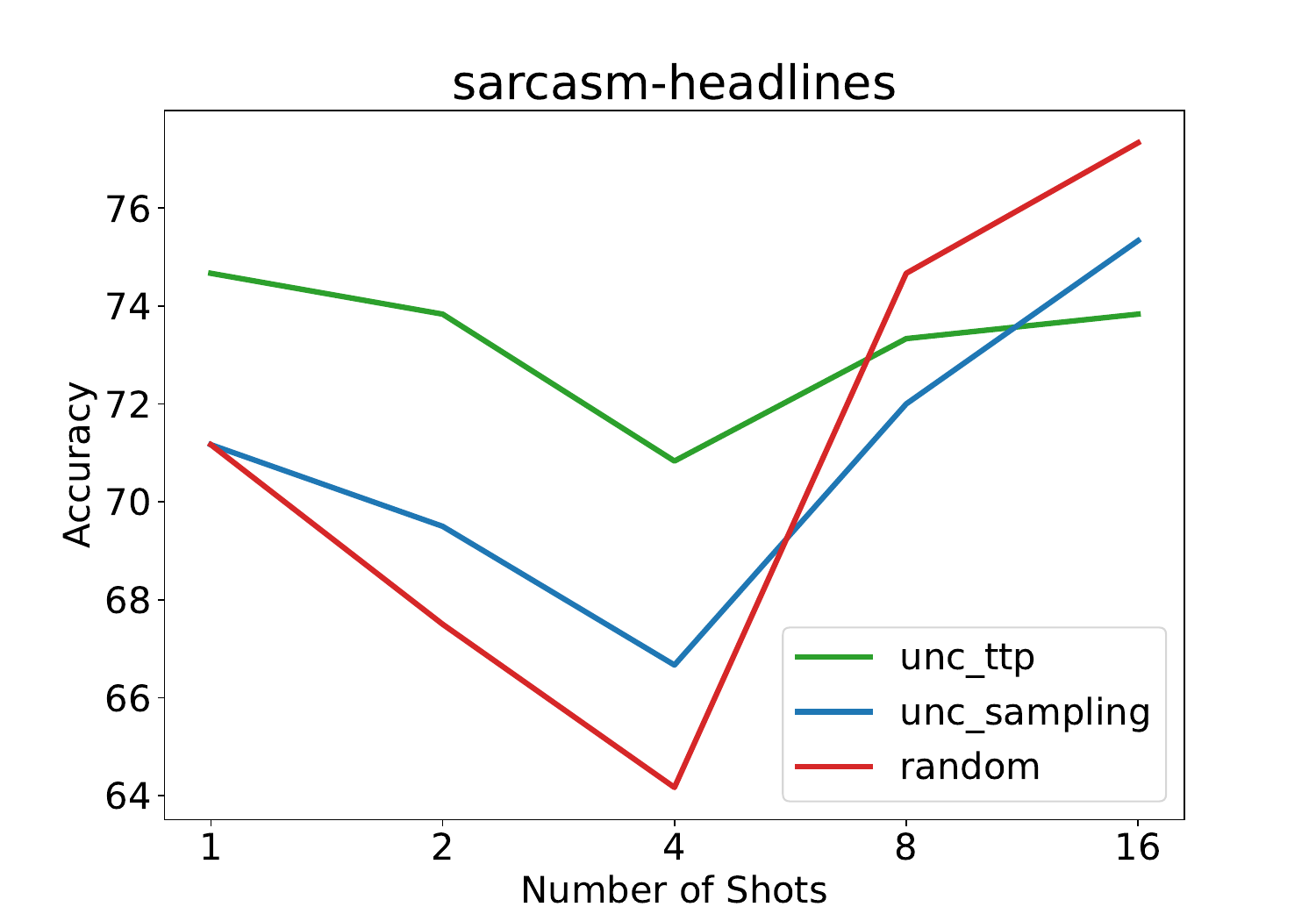}\hspace{0em}
  \includegraphics[width=0.25\linewidth]{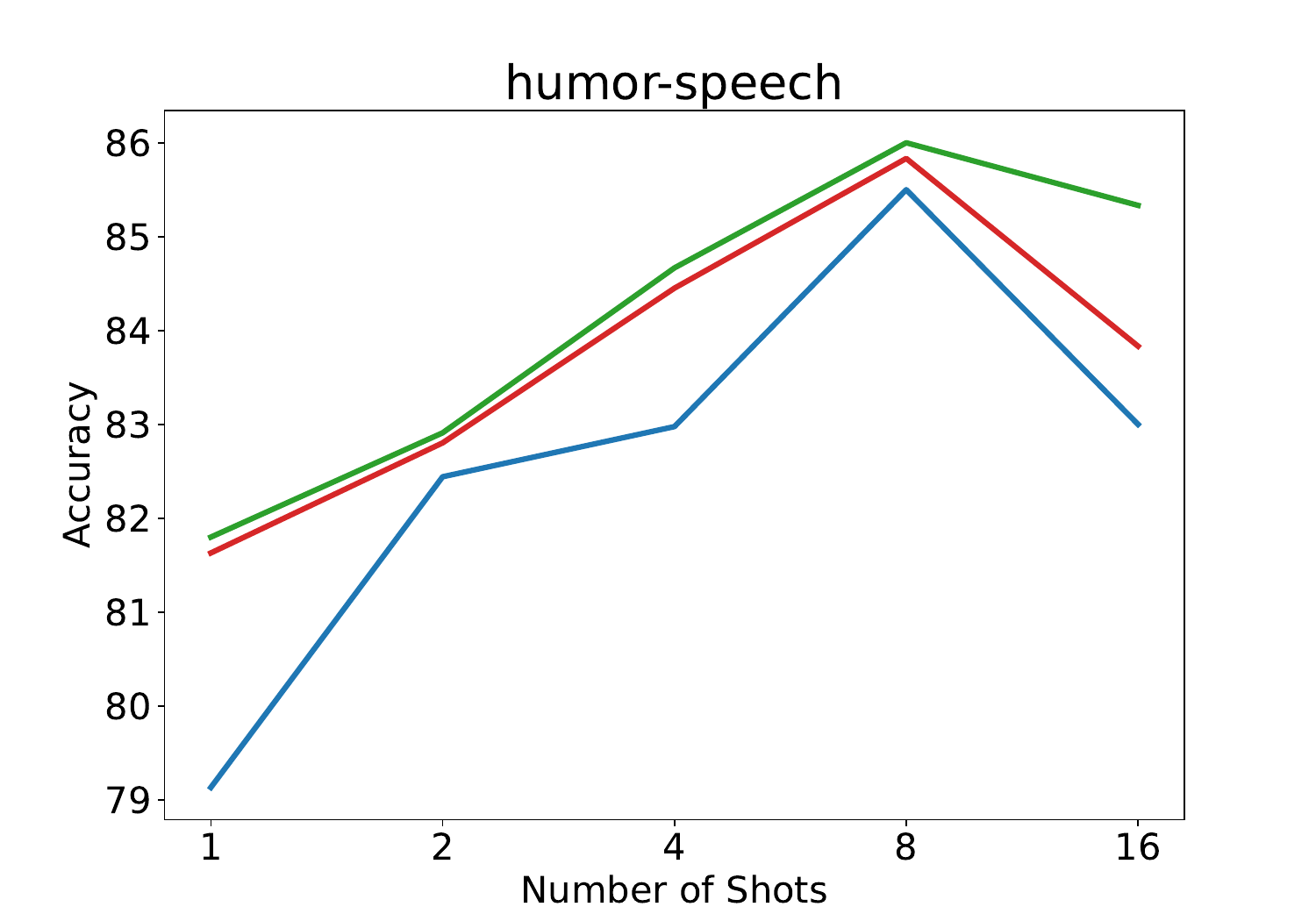}\hspace{0em}
  \includegraphics[width=0.25\linewidth]{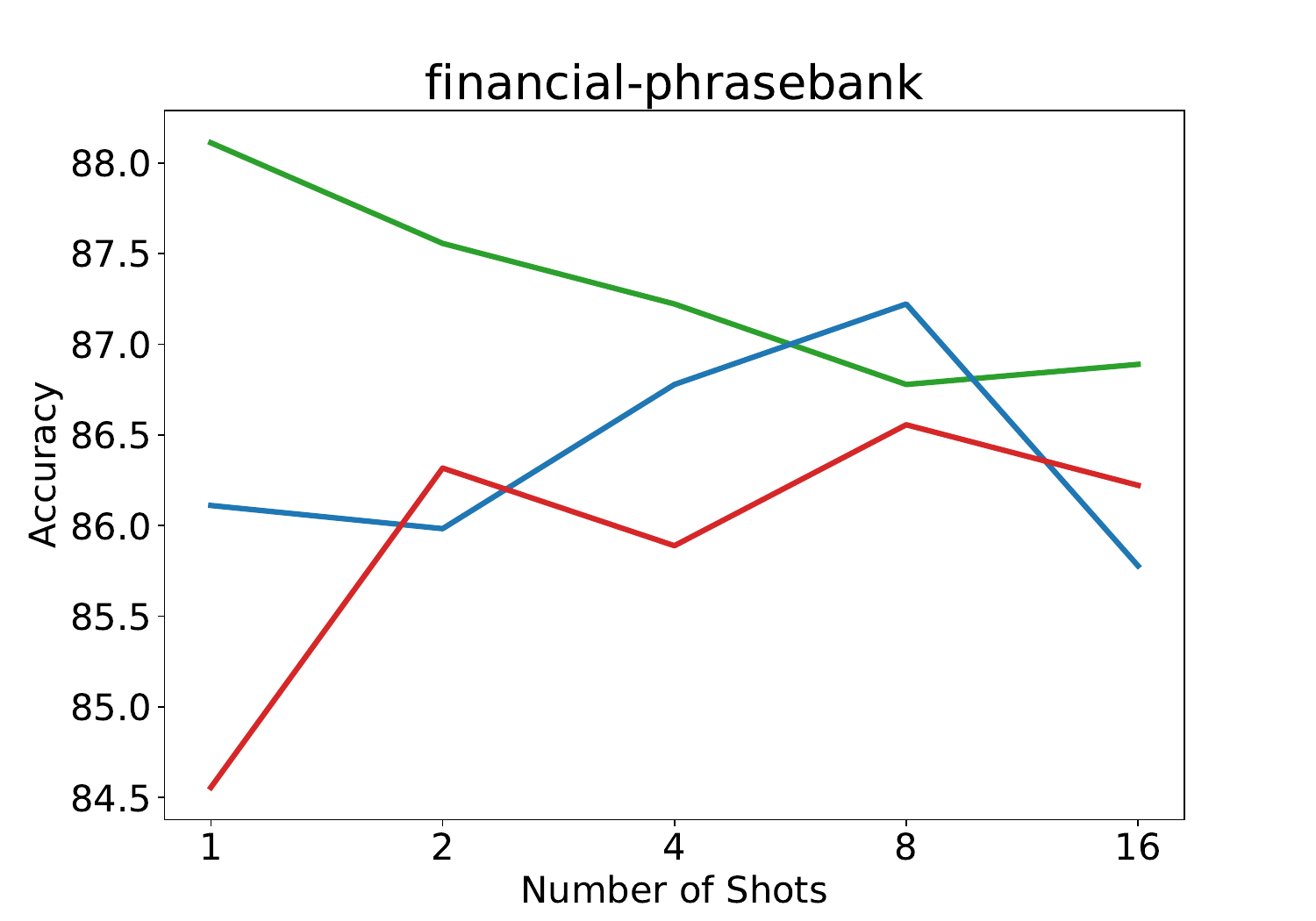}\hspace{0em}
  \caption {Accuracy scores of Mistral $K$-way $N$-shot experiments on the test set. Our Unc-TTP robustly surpasses the random baseline and the sampling-based method when the shot number is small.}
  \label{fig:mistral_line}
\end{figure*}

To explore the informativeness of the selected inconsistency-defined uncertain in-context examples as the number of shot increases, we conduct $N$-shot experiments on Mistral, taking the AL baseline method  \textit{Random} as comparison.

The results, depicted in Figure~\ref{fig:mistral_line}, reveal that uncertainty examples identified by Unc-TTP are generally more informative than those selected by the vanilla sampling-based method or chosen randomly. Unc-TTP consistently outperforms other baselines across 1 to 4 shots, highlighting the informativeness of the examples selected under Unc-TTP's guidance. Although the advantages of inconsistency-defined uncertainty sampling diminish for the SH and FP datasets when the number of shots exceeds 8, we argue that this result is reasonable because as the number of shots increases, it is impossible to avoid introducing other variables, such as diversity and length of the instances, and the current exploration of the best combination of in-context examples is still ongoing, which will also be the direction of our future work. 


\subsection{Assessing the Transferability of Unc-TTP}
\label{sec:transferability}
We have shown that the wavering behavior of LLMs under limited sampling reflects their intrinsic uncertainty, and we leverage this inconsistency-defined uncertainty for active in-context example selection. In Section \ref{sec:motivation}, we observed that more capable LLMs better resist skepticism or incorrect information, maintaining their stance while achieving higher accuracy. Based on this, we explore whether Unc-TTP selects more informative uncertain instances—where stronger LLMs waver—that provide stronger guidance for weaker LLMs compared to their self-guided instances.

We implemented Unc-TTP using GPT-4 and GPT-3.5, selecting examples based on the uncertainty guidance of these models as demonstrations for Llama-2 and Mistral (denoted as Unc${GPT4}$ and Unc${GPT3.5}$). The results in Table~\ref{tab:gpt4-2-weakmodel} show that both strong and weak model uncertainty examples lead to a greater performance gain compared to the \textit{Random} baseline. 
Additionally, the weak model's accuracy with self-guided uncertainty examples surpasses that with uncertainty examples guided by the strong model.
This suggests that Unc-TTP, whether applied to strong or weak models, selects examples that are universally challenging for all LLMs. In particular, the self-guided Unc-TTP helps the model identify model-specific uncertainty samples, further demonstrating its ability to reveal the intrinsic uncertainty inherent to each LLM.


\begin{table}[!t]
    \centering
    \small
    \setlength{\tabcolsep}{1.3mm} 
    \begin{tabular}{lcccc}
    \toprule
        \textbf{Method} & \textbf{SH} & \textbf{HS} & \textbf{FP} & \textbf{Average} \\
        \midrule
        \multicolumn{4}{l}{\textbf{Llama-2}} \\ [3pt]
        Random & 69.8 (1.6) & 53.5 (1.5) & 78.3 (1.8) & 67.2 (1.6) \\
        Unc$_{GPT3.5}$ & 67.8 (1.5) & \textbf{58.8} (1.6) & \textbf{81.2} (2.2) & 69.3 (1.8) \\
        Unc$_{GPT4}$   & 71.0 (2.1) & 56.8 (2.6) & 80.7 (1.5) & 69.5 (2.1) \\
        \cellcolor{gray!20}Unc$_{Llama2}$ & \cellcolor{gray!20}\textbf{71.2} (2.1) & \cellcolor{gray!20}\textbf{58.8} (3.2) & \cellcolor{gray!20}80.8 (1.3) & \cellcolor{gray!20}\textbf{70.3} (2.2) \\
        \midrule [1pt] 
        \multicolumn{4}{l}{\textbf{Mistral}} \\ [3pt]
        Random & 71.2 (1.6) & 81.6 (4.4) & 84.6 (2.3) & 79.1 (2.8) \\
        Unc$_{GPT3.5}$ & 71.7 (3.6) & 79.6 (1.6) & 86.9 (2.0) & 79.4 (3.3) \\
        Unc$_{GPT4}$   & \textbf{75.0} (2.0) & 77.1 (4.9) & 85.8 (2.4) & 79.3 (3.1) \\
        \cellcolor{gray!20}Unc$_{Mistral}$ & \cellcolor{gray!20}74.7 (0.8) & \cellcolor{gray!20}\textbf{81.8} (2.9) & \cellcolor{gray!20}\textbf{88.1} (0.8) & \cellcolor{gray!20}\textbf{81.5} (1.5) \\
        \bottomrule
    \end{tabular}
    \caption{Transferability of Unc-TTP-guided uncertainty sampling in 1-shot ICL: Comparing the informativeness of uncertainty examples selected by strong models versus those selected through weak model self-guidance.}
\label{tab:gpt4-2-weakmodel}
\end{table}


\section{Conclusion}
\label{sec:conclusion}
In this paper, we introduce the Unc-TTP, a novel method for assessing intrinsic uncertainty in both open- and closed-source LLMs. From the perspective of applying active learning principles to in-context example selection, we experimentally demonstrate that uncertainty examples selected through Unc-TTP are more informative than certainty examples. This provides further evidence that output inconsistency can serve as an indicator of LLMs' intrinsic uncertainty. 

Additionally, we propose an uncertainty-based active in-context example selection strategy grounded in Unc-TTP, which outperforms previous AL strategies in terms of performance gains in ICL. These findings establish Unc-TTP as an effective tool for classifying LLM uncertainty and enhancing ICL performance, paving the way for understanding the uncertainty behavior of LLMs.



\section*{Limitations}
Firstly, our proposed Unc-TTP inherits the prompt-sensitive nature of the LLMs, which may pose a risk when applying Unc-TTP-guided uncertainty sampling. However, for the robustness of our experiments, we repeated our experiments three times and reported the mean and standard deviation.

Secondly, we have only evaluated Unc-TTP on subjective tasks. Given highly subjective tasks often require a deep understanding of contextual subtleties, their labels are sometimes "biased" themselves depending on the annotator decomposition~\citep{10.1145/3514094.3534142, li_synthetic_2023}. This characteristic is more compatible with Unc-TTP, as it can effectively expose areas where LLMs exhibit uncertainty compared to objective reasoning tasks. For the implementation of Unc-TTP on objective tasks will be a primary focus in our future work.

Thirdly, we have only tested Unc-TTP on classification tasks, as designing intuitive perturbations, like label injection, for generative or more complex tasks is challenging. This is because such perturbations inevitably introduce additional variables. One possible approach is to transform the LLM’s output into a binary fact-checking task (e.g., SelfCheckGPT). However, our paper shows that this verification-based method is ineffective under the scenario of in-context example selection. Furthermore, Unc-TTP is specifically designed for active context example selection, whereas complex generation tasks and ICL may not be compatible. Developing appropriate perturbations to extend Unc-TTP to more general generative tasks will be the focus of our future work.

Finally, although Unc-TTP adopts a one-example-for-all approach, which makes inference more efficient compared to retrieval-based methods like similarity, it requires more resources for sample selection. However, it is important to note that Unc-TTP not only serves as a novel active example selection method but also as an uncertainty classification method that can reflect the LLM's intrinsic uncertainty. In the future, we aim to explore additional application scenarios for Unc-TTP, a perturbation-based, inconsistency-defined uncertainty classification method.


\section*{Ethics Statement}
\begin{table}[!htbp]
  \centering
  \begin{tabular}{ll}
    \toprule 
    \textbf{Task} & \textbf{Time (min)}\\
    \midrule
    Unc-TTP on Train Set & 30 \\
    1-shot ICL on Validation Set & 100 \\
    1-shot ICL on Test Set & 10 \\
    \bottomrule
  \end{tabular}
  \caption{The approximate computation time of the experiments.
  \textit{Unc-TTP on Train Set} represents the classification of uncertainty categories using Unc-TTP. \textit{1-shot ICL on Validation Set} represents the evaluation of 1-shot examples from all Unc-TTP categories. \textit{1-shot ICL on Test Set} represents the application of selected uncertainty examples on the test set.}
  \label{tab:time}
\end{table}
\paragraph{Computational Budget}
We run all our experiments on Intel$^\circledR$ Xeon$^\circledR$ Silver 4310 CPU and A40 with a batch size of 1 and the max token number of 20. The Table~\ref{tab:time} shows the approximate average computation time with the 7B model Llama-2 and Mistral as the LLM, using 1 random seed.

\paragraph{Use of AI Assistants}
We have employed ChatGPT as a writing assistant, primarily for polishing the text after the initial composition.


\bibliography{acl}

\begin{thebibliography}{41}
\providecommand{\natexlab}[1]{#1}

\bibitem[{Alkaissi and McFarlane(2023)}]{alkaissi2023artificial}
Hussam Alkaissi and Samy~I McFarlane. 2023.
\newblock Artificial hallucinations in chatgpt: implications in scientific writing.
\newblock \emph{Cureus}, 15(2).

\bibitem[{Annamoradnejad and Zoghi(2020)}]{Annamoradnejad_2024}
Issa Annamoradnejad and Gohar Zoghi. 2020.
\newblock \href {https://doi.org/10.1016/j.eswa.2024.123685} {Colbert: Using bert sentence embedding in parallel neural networks for computational humor}.
\newblock \emph{Expert Systems with Applications}, 249:123685.

\bibitem[{Brown et~al.(2020)Brown, Mann, Ryder, Subbiah, Kaplan, Dhariwal, Neelakantan, Shyam, Sastry, Askell, Agarwal, Herbert-Voss, Krueger, Henighan, Child, Ramesh, Ziegler, Wu, Winter, Hesse, Chen, Sigler, Litwin, Gray, Chess, Clark, Berner, McCandlish, Radford, Sutskever, and Amodei}]{Few-Shot_Learners}
Tom Brown, Benjamin Mann, Nick Ryder, Melanie Subbiah, Jared~D Kaplan, Prafulla Dhariwal, Arvind Neelakantan, Pranav Shyam, Girish Sastry, Amanda Askell, Sandhini Agarwal, Ariel Herbert-Voss, Gretchen Krueger, Tom Henighan, Rewon Child, Aditya Ramesh, Daniel Ziegler, Jeffrey Wu, Clemens Winter, Chris Hesse, Mark Chen, Eric Sigler, Mateusz Litwin, Scott Gray, Benjamin Chess, Jack Clark, Christopher Berner, Sam McCandlish, Alec Radford, Ilya Sutskever, and Dario Amodei. 2020.
\newblock \href {https://proceedings.neurips.cc/paper_files/paper/2020/file/1457c0d6bfcb4967418bfb8ac142f64a-Paper.pdf} {Language models are few-shot learners}.
\newblock In \emph{Advances in Neural Information Processing Systems}, volume~33, pages 1877--1901. Curran Associates, Inc.

\bibitem[{Dong et~al.(2023)Dong, Li, Dai, Zheng, Wu, Chang, Sun, Xu, Li, and Sui}]{dong2023survey}
Qingxiu Dong, Lei Li, Damai Dai, Ce~Zheng, Zhiyong Wu, Baobao Chang, Xu~Sun, Jingjing Xu, Lei Li, and Zhifang Sui. 2023.
\newblock \href {https://arxiv.org/abs/2301.00234} {A survey on in-context learning}.
\newblock \emph{Preprint}, arXiv:2301.00234.

\bibitem[{Gonen et~al.(2024)Gonen, Iyer, Blevins, Smith, and Zettlemoyer}]{gonen2024demystifyingpromptslanguagemodels}
Hila Gonen, Srini Iyer, Terra Blevins, Noah~A. Smith, and Luke Zettlemoyer. 2024.
\newblock \href {https://arxiv.org/abs/2212.04037} {Demystifying prompts in language models via perplexity estimation}.
\newblock \emph{Preprint}, arXiv:2212.04037.

\bibitem[{Ji et~al.(2023)Ji, Lee, Frieske, Yu, Su, Xu, Ishii, Bang, Madotto, and Fung}]{Survey_Hallucination}
Ziwei Ji, Nayeon Lee, Rita Frieske, Tiezheng Yu, Dan Su, Yan Xu, Etsuko Ishii, Ye~Jin Bang, Andrea Madotto, and Pascale Fung. 2023.
\newblock \href {https://doi.org/10.1145/3571730} {Survey of hallucination in natural language generation}.
\newblock \emph{ACM Comput. Surv.}, 55(12).

\bibitem[{Jiang et~al.(2023)Jiang, Sablayrolles, Mensch, Bamford, Chaplot, de~las Casas, Bressand, Lengyel, Lample, Saulnier, Lavaud, Lachaux, Stock, Scao, Lavril, Wang, Lacroix, and Sayed}]{jiang2023mistral}
Albert~Q. Jiang, Alexandre Sablayrolles, Arthur Mensch, Chris Bamford, Devendra~Singh Chaplot, Diego de~las Casas, Florian Bressand, Gianna Lengyel, Guillaume Lample, Lucile Saulnier, Lélio~Renard Lavaud, Marie-Anne Lachaux, Pierre Stock, Teven~Le Scao, Thibaut Lavril, Thomas Wang, Timothée Lacroix, and William~El Sayed. 2023.
\newblock \href {https://arxiv.org/abs/2310.06825} {Mistral 7b}.
\newblock \emph{Preprint}, arXiv:2310.06825.

\bibitem[{Kadavath et~al.(2022)Kadavath, Conerly, Askell, Henighan, Drain, Perez, Schiefer, Hatfield-Dodds, DasSarma, Tran-Johnson, Johnston, El-Showk, Jones, Elhage, Hume, Chen, Bai, Bowman, Fort, Ganguli, Hernandez, Jacobson, Kernion, Kravec, Lovitt, Ndousse, Olsson, Ringer, Amodei, Brown, Clark, Joseph, Mann, McCandlish, Olah, and Kaplan}]{kadavath2022languagemodelsmostlyknow}
Saurav Kadavath, Tom Conerly, Amanda Askell, Tom Henighan, Dawn Drain, Ethan Perez, Nicholas Schiefer, Zac Hatfield-Dodds, Nova DasSarma, Eli Tran-Johnson, Scott Johnston, Sheer El-Showk, Andy Jones, Nelson Elhage, Tristan Hume, Anna Chen, Yuntao Bai, Sam Bowman, Stanislav Fort, Deep Ganguli, Danny Hernandez, Josh Jacobson, Jackson Kernion, Shauna Kravec, Liane Lovitt, Kamal Ndousse, Catherine Olsson, Sam Ringer, Dario Amodei, Tom Brown, Jack Clark, Nicholas Joseph, Ben Mann, Sam McCandlish, Chris Olah, and Jared Kaplan. 2022.
\newblock \href {https://arxiv.org/abs/2207.05221} {Language models (mostly) know what they know}.
\newblock \emph{Preprint}, arXiv:2207.05221.

\bibitem[{Krishna et~al.(2023)Krishna, Ma, Slack, Ghandeharioun, Singh, and Lakkaraju}]{Post_Hoc_Explanations}
Satyapriya Krishna, Jiaqi Ma, Dylan Slack, Asma Ghandeharioun, Sameer Singh, and Himabindu Lakkaraju. 2023.
\newblock \href {https://proceedings.neurips.cc/paper_files/paper/2023/file/ce65173b994cf7c925c71b482ee14a8d-Paper-Conference.pdf} {Post hoc explanations of language models can improve language models}.
\newblock In \emph{Advances in Neural Information Processing Systems}, volume~36, pages 65468--65483. Curran Associates, Inc.

\bibitem[{Kumar and Talukdar(2021)}]{kumar2021reordering}
Sawan Kumar and Partha Talukdar. 2021.
\newblock \href {https://arxiv.org/abs/2106.01751} {Reordering examples helps during priming-based few-shot learning}.
\newblock \emph{Preprint}, arXiv:2106.01751.

\bibitem[{Lewis and Catlett(1994)}]{LEWIS1994148}
David~D. Lewis and Jason Catlett. 1994.
\newblock \href {https://doi.org/10.1016/B978-1-55860-335-6.50026-X} {Heterogeneous uncertainty sampling for supervised learning}.
\newblock In William~W. Cohen and Haym Hirsh, editors, \emph{Machine Learning Proceedings 1994}, pages 148--156. Morgan Kaufmann, San Francisco (CA).

\bibitem[{Lewis and Gale(1994)}]{lewis1995sequential}
David~D. Lewis and William~A. Gale. 1994.
\newblock A sequential algorithm for training text classifiers.
\newblock In \emph{Proceedings of the 17th Annual International ACM SIGIR Conference on Research and Development in Information Retrieval}, SIGIR '94, page 3–12, Berlin, Heidelberg. Springer-Verlag.

\bibitem[{Li et~al.(2022)Li, Lu, and Yin}]{10.1145/3514094.3534142}
Zhuoyan Li, Zhuoran Lu, and Ming Yin. 2022.
\newblock \href {https://doi.org/10.1145/3514094.3534142} {Towards better detection of biased language with scarce, noisy, and biased annotations}.
\newblock In \emph{Proceedings of the 2022 AAAI/ACM Conference on AI, Ethics, and Society}, AIES '22, page 411–423, New York, NY, USA. Association for Computing Machinery.

\bibitem[{Li et~al.(2023)Li, Zhu, Lu, and Yin}]{li_synthetic_2023}
Zhuoyan Li, Hangxiao Zhu, Zhuoran Lu, and Ming Yin. 2023.
\newblock \href {https://doi.org/10.18653/v1/2023.emnlp-main.647} {Synthetic data generation with large language models for text classification: Potential and limitations}.
\newblock In \emph{Proceedings of the 2023 Conference on Empirical Methods in Natural Language Processing}, pages 10443--10461. Association for Computational Linguistics.

\bibitem[{Liu et~al.(2021)Liu, Shen, Zhang, Dolan, Carin, and Chen}]{liu2021makes}
Jiachang Liu, Dinghan Shen, Yizhe Zhang, Bill Dolan, Lawrence Carin, and Weizhu Chen. 2021.
\newblock \href {https://arxiv.org/abs/2101.06804} {What makes good in-context examples for gpt-$3$?}
\newblock \emph{Preprint}, arXiv:2101.06804.

\bibitem[{Lu et~al.(2022)Lu, Bartolo, Moore, Riedel, and Stenetorp}]{lu2022fantastically}
Yao Lu, Max Bartolo, Alastair Moore, Sebastian Riedel, and Pontus Stenetorp. 2022.
\newblock \href {https://arxiv.org/abs/2104.08786} {Fantastically ordered prompts and where to find them: Overcoming few-shot prompt order sensitivity}.
\newblock \emph{Preprint}, arXiv:2104.08786.

\bibitem[{Malo et~al.(2014)Malo, Sinha, Korhonen, Wallenius, and Takala}]{malo_good_2014}
Pekka Malo, Ankur Sinha, Pekka Korhonen, Jyrki Wallenius, and Pyry Takala. 2014.
\newblock \href {https://doi.org/10.1002/asi.23062} {Good debt or bad debt: Detecting semantic orientations in economic texts}.
\newblock \emph{Journal of the Association for Information Science and Technology}, 65(4):782--796.

\bibitem[{Manakul et~al.(2023)Manakul, Liusie, and Gales}]{manakul-etal-2023-selfcheckgpt}
Potsawee Manakul, Adian Liusie, and Mark Gales. 2023.
\newblock \href {https://doi.org/10.18653/v1/2023.emnlp-main.557} {{S}elf{C}heck{GPT}: Zero-resource black-box hallucination detection for generative large language models}.
\newblock In \emph{Proceedings of the 2023 Conference on Empirical Methods in Natural Language Processing}, pages 9004--9017, Singapore. Association for Computational Linguistics.

\bibitem[{Margatina et~al.(2023)Margatina, Schick, Aletras, and Dwivedi-Yu}]{margatina2023activelearningprinciplesincontext}
Katerina Margatina, Timo Schick, Nikolaos Aletras, and Jane Dwivedi-Yu. 2023.
\newblock \href {https://arxiv.org/abs/2305.14264} {Active learning principles for in-context learning with large language models}.
\newblock \emph{Preprint}, arXiv:2305.14264.

\bibitem[{Maynez et~al.(2020)Maynez, Narayan, Bohnet, and McDonald}]{maynez2020faithfulness}
Joshua Maynez, Shashi Narayan, Bernd Bohnet, and Ryan McDonald. 2020.
\newblock \href {https://arxiv.org/abs/2005.00661} {On faithfulness and factuality in abstractive summarization}.
\newblock \emph{Preprint}, arXiv:2005.00661.

\bibitem[{Misra and Arora(2023)}]{misra_sarcasm_2023}
Rishabh Misra and Prahal Arora. 2023.
\newblock \href {https://doi.org/10.1016/j.aiopen.2023.01.001} {Sarcasm detection using news headlines dataset}.
\newblock \emph{{AI} Open}, 4:13--18.

\bibitem[{Mo et~al.(2024)Mo, Liu, Yang, Wang, Zhang, Wang, and Li}]{mo2024ciclcontrastiveincontextlearning}
Ying Mo, Jiahao Liu, Jian Yang, Qifan Wang, Shun Zhang, Jingang Wang, and Zhoujun Li. 2024.
\newblock \href {https://arxiv.org/abs/2402.11254} {C-icl: Contrastive in-context learning for information extraction}.
\newblock \emph{Preprint}, arXiv:2402.11254.

\bibitem[{OpenAI(2024)}]{openai2024gpt4}
OpenAI. 2024.
\newblock \href {https://arxiv.org/abs/2303.08774} {Gpt-4 technical report}.
\newblock \emph{Preprint}, arXiv:2303.08774.

\bibitem[{Perez et~al.(2022)Perez, Ringer, Lukošiūtė, Nguyen, Chen, Heiner, Pettit, Olsson, Kundu, Kadavath, Jones, Chen, Mann, Israel, Seethor, McKinnon, Olah, Yan, Amodei, Amodei, Drain, Li, Tran-Johnson, Khundadze, Kernion, Landis, Kerr, Mueller, Hyun, Landau, Ndousse, Goldberg, Lovitt, Lucas, Sellitto, Zhang, Kingsland, Elhage, Joseph, Mercado, DasSarma, Rausch, Larson, McCandlish, Johnston, Kravec, Showk, Lanham, Telleen-Lawton, Brown, Henighan, Hume, Bai, Hatfield-Dodds, Clark, Bowman, Askell, Grosse, Hernandez, Ganguli, Hubinger, Schiefer, and Kaplan}]{perez2022discoveringlanguagemodelbehaviors}
Ethan Perez, Sam Ringer, Kamilė Lukošiūtė, Karina Nguyen, Edwin Chen, Scott Heiner, Craig Pettit, Catherine Olsson, Sandipan Kundu, Saurav Kadavath, Andy Jones, Anna Chen, Ben Mann, Brian Israel, Bryan Seethor, Cameron McKinnon, Christopher Olah, Da~Yan, Daniela Amodei, Dario Amodei, Dawn Drain, Dustin Li, Eli Tran-Johnson, Guro Khundadze, Jackson Kernion, James Landis, Jamie Kerr, Jared Mueller, Jeeyoon Hyun, Joshua Landau, Kamal Ndousse, Landon Goldberg, Liane Lovitt, Martin Lucas, Michael Sellitto, Miranda Zhang, Neerav Kingsland, Nelson Elhage, Nicholas Joseph, Noemí Mercado, Nova DasSarma, Oliver Rausch, Robin Larson, Sam McCandlish, Scott Johnston, Shauna Kravec, Sheer~El Showk, Tamera Lanham, Timothy Telleen-Lawton, Tom Brown, Tom Henighan, Tristan Hume, Yuntao Bai, Zac Hatfield-Dodds, Jack Clark, Samuel~R. Bowman, Amanda Askell, Roger Grosse, Danny Hernandez, Deep Ganguli, Evan Hubinger, Nicholas Schiefer, and Jared Kaplan. 2022.
\newblock \href {https://arxiv.org/abs/2212.09251} {Discovering language model behaviors with model-written evaluations}.
\newblock \emph{Preprint}, arXiv:2212.09251.

\bibitem[{Reimers and Gurevych(2019)}]{2019sentencebert}
Nils Reimers and Iryna Gurevych. 2019.
\newblock \href {https://arxiv.org/abs/1908.10084} {Sentence-bert: Sentence embeddings using siamese bert-networks}.
\newblock \emph{Preprint}, arXiv:1908.10084.

\bibitem[{Robertson and Zaragoza(2009)}]{INR-019}
Stephen Robertson and Hugo Zaragoza. 2009.
\newblock \href {https://doi.org/10.1561/1500000019} {The probabilistic relevance framework: Bm25 and beyond}.
\newblock \emph{Foundations and Trends® in Information Retrieval}, 3(4):333--389.

\bibitem[{Settles(2009)}]{settles_active_2009}
Burr Settles. 2009.
\newblock \href {https://minds.wisconsin.edu/handle/1793/60660} {Active learning literature survey}.

\bibitem[{Sharma et~al.(2023)Sharma, Tong, Korbak, Duvenaud, Askell, Bowman, Cheng, Durmus, Hatfield-Dodds, Johnston, Kravec, Maxwell, McCandlish, Ndousse, Rausch, Schiefer, Yan, Zhang, and Perez}]{sharma2023understanding}
Mrinank Sharma, Meg Tong, Tomasz Korbak, David Duvenaud, Amanda Askell, Samuel~R. Bowman, Newton Cheng, Esin Durmus, Zac Hatfield-Dodds, Scott~R. Johnston, Shauna Kravec, Timothy Maxwell, Sam McCandlish, Kamal Ndousse, Oliver Rausch, Nicholas Schiefer, Da~Yan, Miranda Zhang, and Ethan Perez. 2023.
\newblock \href {https://arxiv.org/abs/2310.13548} {Towards understanding sycophancy in language models}.
\newblock \emph{Preprint}, arXiv:2310.13548.

\bibitem[{Shen et~al.(2023)Shen, Chen, Backes, and Zhang}]{shen2023chatgpt}
Xinyue Shen, Zeyuan Chen, Michael Backes, and Yang Zhang. 2023.
\newblock \href {https://arxiv.org/abs/2304.08979} {In chatgpt we trust? measuring and characterizing the reliability of chatgpt}.
\newblock \emph{Preprint}, arXiv:2304.08979.

\bibitem[{Touvron et~al.(2023)Touvron, Martin, Stone, Albert, Almahairi, Babaei, Bashlykov, Batra, Bhargava, Bhosale, Bikel, Blecher, Ferrer, Chen, Cucurull, Esiobu, Fernandes, Fu, Fu, Fuller, Gao, Goswami, Goyal, Hartshorn, Hosseini, Hou, Inan, Kardas, Kerkez, Khabsa, Kloumann, Korenev, Koura, Lachaux, Lavril, Lee, Liskovich, Lu, Mao, Martinet, Mihaylov, Mishra, Molybog, Nie, Poulton, Reizenstein, Rungta, Saladi, Schelten, Silva, Smith, Subramanian, Tan, Tang, Taylor, Williams, Kuan, Xu, Yan, Zarov, Zhang, Fan, Kambadur, Narang, Rodriguez, Stojnic, Edunov, and Scialom}]{touvron2023llama}
Hugo Touvron, Louis Martin, Kevin Stone, Peter Albert, Amjad Almahairi, Yasmine Babaei, Nikolay Bashlykov, Soumya Batra, Prajjwal Bhargava, Shruti Bhosale, Dan Bikel, Lukas Blecher, Cristian~Canton Ferrer, Moya Chen, Guillem Cucurull, David Esiobu, Jude Fernandes, Jeremy Fu, Wenyin Fu, Brian Fuller, Cynthia Gao, Vedanuj Goswami, Naman Goyal, Anthony Hartshorn, Saghar Hosseini, Rui Hou, Hakan Inan, Marcin Kardas, Viktor Kerkez, Madian Khabsa, Isabel Kloumann, Artem Korenev, Punit~Singh Koura, Marie-Anne Lachaux, Thibaut Lavril, Jenya Lee, Diana Liskovich, Yinghai Lu, Yuning Mao, Xavier Martinet, Todor Mihaylov, Pushkar Mishra, Igor Molybog, Yixin Nie, Andrew Poulton, Jeremy Reizenstein, Rashi Rungta, Kalyan Saladi, Alan Schelten, Ruan Silva, Eric~Michael Smith, Ranjan Subramanian, Xiaoqing~Ellen Tan, Binh Tang, Ross Taylor, Adina Williams, Jian~Xiang Kuan, Puxin Xu, Zheng Yan, Iliyan Zarov, Yuchen Zhang, Angela Fan, Melanie Kambadur, Sharan Narang, Aurelien Rodriguez, Robert Stojnic, Sergey Edunov, and Thomas
  Scialom. 2023.
\newblock \href {https://arxiv.org/abs/2307.09288} {Llama 2: Open foundation and fine-tuned chat models}.
\newblock \emph{Preprint}, arXiv:2307.09288.

\bibitem[{Turpin et~al.(2023{\natexlab{a}})Turpin, Michael, Perez, and Bowman}]{turpin_language_2023}
Miles Turpin, Julian Michael, Ethan Perez, and Samuel Bowman. 2023{\natexlab{a}}.
\newblock \href {https://proceedings.neurips.cc/paper_files/paper/2023/file/ed3fea9033a80fea1376299fa7863f4a-Paper-Conference.pdf} {Language models don' t always say what they think: Unfaithful explanations in chain-of-thought prompting}.
\newblock In \emph{Advances in Neural Information Processing Systems}, volume~36, pages 74952--74965. Curran Associates, Inc.

\bibitem[{Turpin et~al.(2023{\natexlab{b}})Turpin, Michael, Perez, and Bowman}]{Unfaithful_CoT}
Miles Turpin, Julian Michael, Ethan Perez, and Samuel Bowman. 2023{\natexlab{b}}.
\newblock \href {https://proceedings.neurips.cc/paper_files/paper/2023/file/ed3fea9033a80fea1376299fa7863f4a-Paper-Conference.pdf} {Language models don\textquotesingle t always say what they think: Unfaithful explanations in chain-of-thought prompting}.
\newblock In \emph{Advances in Neural Information Processing Systems}, volume~36, pages 74952--74965. Curran Associates, Inc.

\bibitem[{Wang et~al.(2023{\natexlab{a}})Wang, Yue, and Sun}]{wang-etal-2023-chatgpt-defend}
Boshi Wang, Xiang Yue, and Huan Sun. 2023{\natexlab{a}}.
\newblock \href {https://doi.org/10.18653/v1/2023.findings-emnlp.795} {Can {C}hat{GPT} defend its belief in truth? evaluating {LLM} reasoning via debate}.
\newblock In \emph{Findings of the Association for Computational Linguistics: EMNLP 2023}, pages 11865--11881, Singapore. Association for Computational Linguistics.

\bibitem[{Wang et~al.(2023{\natexlab{b}})Wang, Wei, Schuurmans, Le, Chi, Narang, Chowdhery, and Zhou}]{wang2023selfconsistency}
Xuezhi Wang, Jason Wei, Dale Schuurmans, Quoc Le, Ed~Chi, Sharan Narang, Aakanksha Chowdhery, and Denny Zhou. 2023{\natexlab{b}}.
\newblock \href {https://arxiv.org/abs/2203.11171} {Self-consistency improves chain of thought reasoning in language models}.
\newblock \emph{Preprint}, arXiv:2203.11171.

\bibitem[{Wei et~al.(2024{\natexlab{a}})Wei, Huang, Lu, Zhou, and Le}]{wei2024simple}
Jerry Wei, Da~Huang, Yifeng Lu, Denny Zhou, and Quoc~V. Le. 2024{\natexlab{a}}.
\newblock \href {https://arxiv.org/abs/2308.03958} {Simple synthetic data reduces sycophancy in large language models}.
\newblock \emph{Preprint}, arXiv:2308.03958.

\bibitem[{Wei et~al.(2024{\natexlab{b}})Wei, Yao, Ton, Guo, Estornell, and Liu}]{wei2024measuring}
Jiaheng Wei, Yuanshun Yao, Jean-Francois Ton, Hongyi Guo, Andrew Estornell, and Yang Liu. 2024{\natexlab{b}}.
\newblock \href {https://arxiv.org/abs/2402.10412} {Measuring and reducing llm hallucination without gold-standard answers via expertise-weighting}.
\newblock \emph{Preprint}, arXiv:2402.10412.

\bibitem[{Xie et~al.(2024)Xie, Wang, Feng, and Xia}]{xie2024ask}
Qiming Xie, Zengzhi Wang, Yi~Feng, and Rui Xia. 2024.
\newblock \href {https://arxiv.org/abs/2310.02174} {Ask again, then fail: Large language models' vacillations in judgement}.
\newblock \emph{Preprint}, arXiv:2310.02174.

\bibitem[{Xu et~al.(2024{\natexlab{a}})Xu, Zhu, Zhang, Ma, Fan, Chen, and Yu}]{xu2024rejection}
Hongshen Xu, Zichen Zhu, Situo Zhang, Da~Ma, Shuai Fan, Lu~Chen, and Kai Yu. 2024{\natexlab{a}}.
\newblock \href {https://arxiv.org/abs/2403.18349} {Rejection improves reliability: Training llms to refuse unknown questions using rl from knowledge feedback}.
\newblock \emph{Preprint}, arXiv:2403.18349.

\bibitem[{Xu et~al.(2024{\natexlab{b}})Xu, Qi, and Xu}]{xu2024preemptive}
Rongwu Xu, Zehan Qi, and Wei Xu. 2024{\natexlab{b}}.
\newblock \href {https://arxiv.org/abs/2405.20902} {Preemptive answer "attacks" on chain-of-thought reasoning}.
\newblock \emph{Preprint}, arXiv:2405.20902.

\bibitem[{Yang et~al.(2023)Yang, Chern, Qiu, Neubig, and Liu}]{yang2023alignment}
Yuqing Yang, Ethan Chern, Xipeng Qiu, Graham Neubig, and Pengfei Liu. 2023.
\newblock \href {https://arxiv.org/abs/2312.07000} {Alignment for honesty}.
\newblock \emph{Preprint}, arXiv:2312.07000.

\bibitem[{Yu et~al.(2023)Yu, Iter, Wang, Xu, Ju, Sanyal, Zhu, Zeng, and Jiang}]{yu2023generateretrievelargelanguage}
Wenhao Yu, Dan Iter, Shuohang Wang, Yichong Xu, Mingxuan Ju, Soumya Sanyal, Chenguang Zhu, Michael Zeng, and Meng Jiang. 2023.
\newblock \href {https://arxiv.org/abs/2209.10063} {Generate rather than retrieve: Large language models are strong context generators}.
\newblock \emph{Preprint}, arXiv:2209.10063.

\end{thebibliography}

\clearpage

\appendix

\section{Detailed Performance of LLMs under Preliminary Test}
\label{sec:detailbehaviour}

The results of preliminary test under three label injection settings on three benchmarks are shown in Figure~\ref{fig:bar_stage1}. LLM shows more compliance with correct labels. For the setting that provides labels, all models except Llama-2 on the SH dataset show gullibility to the given labels. Particularly, in the \textit{right-label} setting, some models nearly reach 100\% agreement with the given label. However, for the \textit{wrong-label} setting, the model exhibits varying levels of resistance to enticement. Besides, the more capable the model is, the better it maintains accuracy in the \textit{wrong-label} setting, although it does not converge to the accuracy achieved in the \textit{no-label} setting. This indicates that unless the LLM is very certain of its answer, it reacts quite prudently when the labels provided deviate considerably from its own answers, as adopting the user's point of view seems to be a safer option.

LLMs tend to make compromises when conservative options are offered, choosing the neutral option to accommodate the user's preference.
In the FP dataset, all models exhibit relatively steady performance under the \textit{wrong-label} setting. A possible explanation for this observation is that FP is a classification dataset with three labels, where the distinctions between \textit{neutral} and \textit{positive} or \textit{negative} are subtle. LLMs inherently tend to favor the more conservative \textit{neutral} answers. When their initial beliefs are challenged in the \textit{wrong-label} setting, they are inclined to prefer the \textit{neutral} option, potentially aligning with user preferences. 
This behavior corresponds to the data distribution in Table~\ref{ans-distribution-on-fp} of the Appendix. 
We observe that most LLMs exhibit a tendency to produce conservative answers in the \textit{no-label} setting. This tendency becomes pronounced in the \textit{wrong-label} setting, where the number of \textit{neutral} answers doubles, further substantiating our conclusion.

\begin{table}[!htbp]
    \centering
    \setlength{\tabcolsep}{1.4mm} 
    \begin{tabular}{lcccc}
    \toprule 
        \textbf{} & \textbf{Positive} & \textbf{Neutral} & \textbf{Negative} & \textbf{Failed} \\
        \midrule
        \multicolumn{5}{l}{\textbf{\textit{no-label}}} \\ [3pt]
        GPT-4 & 78 & 133 & 89 & 0 \\
        GPT-3.5 & 89 & 138 & 73 & 0 \\
        Mistral & 41 & 179 & 80 & 0 \\
        Llama-2 & 120 & 6 & 174 & 0 \\
        \midrule
        \multicolumn{5}{l}{\textbf{\textit{right-label}}} \\ [3pt]
        GPT-4 & 99 & 101 & 100 & 0 \\
        GPT-3.5 & 97 & 110 & 93 & 0 \\
        Mistral & 84 & 142 & 92 & 0 \\
        Llama-2 & 188 & 9 & 103 & 0 \\
        \midrule
        \multicolumn{5}{l}{\textbf{\textit{wrong-label}}} \\ [3pt]
        GPT-4 & 17 & 249 & 34 & 0 \\
        GPT-3.5 & 42 & 205 & 53 & 0 \\
        Mistral & 0 & 280 & 2 & 18 \\
        Llama-2 & 82 & 43 & 175 & 0 \\
        \bottomrule
    \end{tabular}
    \caption{
    Answer distribution of \textit{no-label} and \textit{wrong-label} setting on Financial Phrasebank (FP), where \textit{Failed} denotes the number of cases LLM fails or rejects to answer. It can be observed that the \textit{wrong-label} doubles the number of neutral responses compared to the \textit{no-label} setting.
  }
  \label{ans-distribution-on-fp}
\end{table}

\section{Experiment Details}
\subsection{Dataset Implementation Details}
\label{sec:dataset-detail}

Since the SH dataset does not provide official data splits, we randomly selected 500 samples as the training set, 1500 as the validation set, and 200 as the test set balancing budget and time constraints. 

For the FP dataset, each sentence is annotated between 5 to 8 times. To ensure high data quality, we included only those sentences with 100\% annotation agreement, resulting in 2,264 examples available for training and evaluation. Among these, only 303 examples were labeled as negative. Consequently, we randomly selected 100 instances for each label to form the training set and 200 instances for each label for the validation set. We randomly sample 200 instances from the split that reach 75\% annotation agreement for the test set. In the \textit{wrong-label} setting of Unc-TTP, for datasets with more than two labels, such as FP, we randomly select an incorrect label to serve as the provided label.

Other data splits are detailed in Table~\ref{tab:data statistics} for reference. It should be noted that all divisions have an equal number of examples for each label. The labels we use in the prompt are as follows:
\begin{itemize}
    \item \textbf{SH}: sarcastic / non-sarcastic
    \item \textbf{HS}: humorous / not humorous
    \item \textbf{FP}: positive / neutral / negative
\end{itemize}

\begin{table}[!htbp]
  \centering
  \begin{tabular}{ccccc}
    \toprule 
    \small
    \textbf{Dataset} & \textbf{Train} & \textbf{Valid} & \textbf{Test} & \textbf{Label} \\
    \midrule
    SH &  500 & 1500 & 200 &  2 \\
    HS &  500 & 1500 & 200 & 2 \\
    FP &  300 & 600 & 300 & 3 \\
    \bottomrule
  \end{tabular}
  \caption{The amount of data and the corresponding number of labels for each dataset we used in our experiments. For all splits, the number of labels is equally distributed.}
  \label{tab:data statistics}
\end{table}

\subsection{Example Selection Details}
\label{sec:exampleSelection-detail}
\paragraph{Uncertainty Category Distribution}
\label{category_distribution}

\begin{figure}[!t]
  \includegraphics[width=\columnwidth]{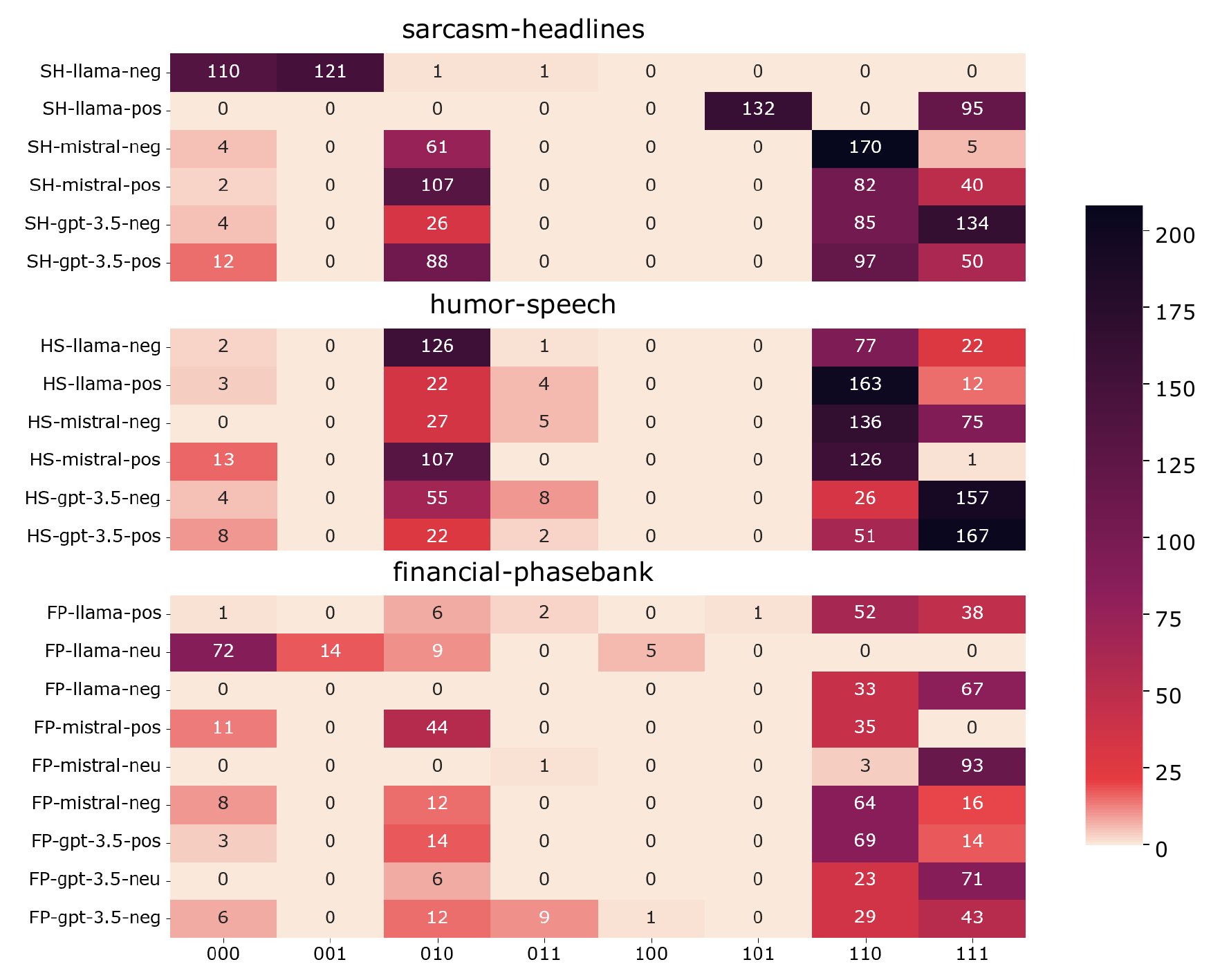}
  \caption{Number of examples corresponding to the respective labels for each model on the three datasets.} 
  \label{fig:heat-map}
\end{figure}

We visualize the respective quantities of instances for each category in Figure~\ref{fig:heat-map}. 
Interestingly, the number of instances in the categories that do not match human intuition, such as \textit{001}, \textit{100}, and \textit{101}, is notably low, as expected.
For instance, if a model misclassifies an instance both without guidance and with correct guidance, it should have a very low likelihood of answering correctly when provided with incorrect labels.

Moreover, the categories we anticipate being the most informative—those that could stimulate deep thinking in the model when presented with a label, such as \textit{011}—are also scarce. This scarcity is likely due to the high level of capability required to generate such categories. The next most informative categories, in our view, are \textit{010} and \textit{110}. These three categories represent the uncertainty types of primary interest to us.

\section{Uncertainty Category Selection}
\label{sec:1-shot-valid} 
We select the final inconsistency-defined uncertainty category of both Unc-TTP and vanilla sampling using validation set due to the category where we regard as informative, such as \textit{011} is scarce, and sometime have not enough data in the particular category to perform $K$-way $N$-shot experiment. 


The Unc-TTP-guided uncertainty sampling result on validation set are shown in the Table \ref{tab:1-shot-valid-uncttp}. For smaller models, the informative uncertainty categories are predominantly centered around 011 and 010. However, for larger models, there is no discernible pattern in the informative uncertainty categories. We speculate that this lack of pattern may be due to two reasons: first, the larger model may rely less on in-context learning (ICL) compared to smaller models; and second, the increased parameter size of the larger model introduces greater stochasticity, even when using greedy decoding.

\begin{table}[!htbp]
    \centering
    \begin{tabular}{@{\extracolsep{\fill}}c|ccc}
    \toprule 
        \textbf{Type} & \textbf{SH} & \textbf{HS} & \textbf{FP} \\
        \midrule
        \multicolumn{4}{l}{\textbf{Llama-2}} \\ [3pt] 
        000 & \underline{63.9} (2.3) & 52.8 (1.8) & 85.5 (0.8) \\
        111 & \underline{63.9} (2.9) & \underline{52.9} (1.4) & \underline{\textbf{87.5}} (1.2) \\
        \hline
        001/010/100 & 62.5 (0.6) & \underline{\textbf{53.8}} (1.1) & 82.1 (3.7) \\
        011/101/110 & \underline{\textbf{64.6}} (0.9) & 53.5 (3.6) & \underline{84.6} (2.8) \\
        \midrule [1pt] 
        \multicolumn{4}{l}{\textbf{Mistral}} \\ [3pt] 
        000 & 65.7 (5.1) & \underline{80.3} (3.1) & 89.2 (2.3) \\
        111 & \underline{71.2} (3.8) & 76.2 (0.7) & \underline{90.0} (0.7) \\
        \hline
        001/010/100 & \underline{\textbf{72.0}} (0.9) & 76.5 (2.1) & \underline{\textbf{90.4}} (0.9) \\
        011/101/110 & 69.2 (4.6) & \underline{\textbf{81.2}} (1.1) & 89.3 (3.6) \\
        \midrule [1pt] 
        \multicolumn{4}{l}{\textbf{GPT-3.5}} \\ [3pt] 
        000 & 71.0 (2.9) & \underline{\textbf{91.9}} (0.5) & \underline{\textbf{91.0}} (1.6) \\
        111 & 76.2 (3.3) & 91.1 (1.2) & 90.5 (1.7) \\
        \hline
        001/010/100 & 73.4 (1.8) & \underline{91.8} (1.9) & \underline{89.8} (2.1) \\
        011/101/110 & \underline{\textbf{77.6}} (3.1) & 90.9 (0.1) & 88.7 (2.8) \\
        \bottomrule
    \end{tabular}
    \caption{Vanilla Sampling method on validation set.}
    \label{tab:1-shot-valid-vanilla}
\end{table}


The utility of Vanilla-Sampling-selected uncertainty examples for each category in the validation set is shown in Table~\ref{tab:1-shot-valid-vanilla}. In vanilla sampling, the number of certainty and uncertainty categories is identical, with each category consisting of two types. For the certainty category, the types are "certainly wrong" (000) and "certainly right" (111). For the uncertainty category, the types are "one right, two wrong" and "one wrong, two right." Therefore, the selection of uncertainty examples is completely balanced. The performance gain observed when uncertainty examples are used in ICL, compared to when certainty examples are used, is not due to any bias in validation set selection.

\begin{table}[!htbp]
    \centering
    \begin{tabular}{@{\extracolsep{\fill}}c|ccc}
    \toprule 
        \textbf{Type} & \textbf{SH} & \textbf{HS} & \textbf{FP} \\ \midrule
        \multicolumn{4}{l}{\textbf{Llama-2}} \\ [3pt] 
        000 & 65.5 (1.6) & \underline{60.3} (5.8) & 86.2 (0.4) \\
        111 & 65.7 (3.1) & 55.6 (2.9) & 83.8 (2.3) \\
        \hline
        001 & 64.4 (1.5) & - & 84.9 (1.6) \\
        010 & \underline{\textbf{68.6}} (0.6) & 58.2 (2.8) & 84.9 (2.6) \\
        011 & \underline{68.0} (1.2) & \underline{\textbf{62.4}} (1.8) & \underline{86.5} (1.9) \\
        100 & - & - & 84.6 (0.9) \\
        101 & 66.5 (2.4) & - & \underline{\textbf{86.6}} (1.4) \\
        110 & - & 57.0 (4.1) & 84.1 (0.9) \\
        \cmidrule{1-4}
        Random & 66.8 (1.7) & 53.4 (3.3) & 84.4 (3.2) \\
        \midrule [1pt] 
        \multicolumn{4}{l}{\textbf{Mistral}} \\ [3pt] 
        000 & 68.2 (3.6) & \underline{81.9} (2.0) & \underline{90.6} (4.0) \\
        111 & 65.4 (2.9) & 75.3 (0.6) & 88.9 (2.1) \\
        \hline
        001 & - & - & - \\
        010 & \underline{\textbf{69.6}} (2.0) & 81.4 (0.9) & \underline{\textbf{91.1}} (1.0) \\
        011 & - & \underline{\textbf{82.0}} (2.6) & 86.3 (0.9) \\
        100 & - & - & - \\
        101 & - & - & - \\
        110 & \underline{68.6} (1.9) & 80.9 (2.9) & 89.9 (1.9) \\
        \cmidrule{1-4}
        Random & 67.3 (1.2) & 81.8 (3.2) & 89.2 (4.3) \\
        \midrule [1pt] 
        \multicolumn{4}{l}{\textbf{ GPT-3.5}} \\ [3pt] 
        000 & 77.0 (2.1) & 88.4 (1.4) & 83.9 (2.1) \\
        111 & \underline{\textbf{81.6}} (2.9) & 89.7 (0.8) & \underline{87.2} (1.5) \\
        \hline
        001 & - & - & - \\
        010 & \underline{77.0} (3.8) & 88.4 (1.5) & 85.3 (1.7) \\
        011 & - & 89.8 (1.0) & 85.6 (2.5) \\
        100 & - & \underline{\textbf{92.3}} (0.3) & 86.9 (1.1) \\
        101 & - & \underline{90.0} (2.3) & - \\
        110 & 76.6 (4.3) & \underline{90.0} (1.4) & \underline{\textbf{88.3}} (2.8) \\
        \cmidrule{1-4}
        Random & 80.4 (2.2) & 89.8 (1.4) & 86.2 (2.6) \\
        \bottomrule
    \end{tabular}
    \caption{1-shot results of LLMs in the validation set. The table presents the mean accuracy across three estimations, with standard deviations in parentheses. The two highest accuracy values under the Unc-TTP setting are underlined. The highest results are highlighted in bold.}
    \label{tab:1-shot-valid-uncttp}
\end{table}

\section{Detailed Results on Accessing the Informative of Inconsistency-Based Methods}

We present the complete results on three datasets for each of the four methods of measuring uncertainty through output inconsistency in Figure \ref{tab:main-unc-cer-test}.

\begin{table}[!htbp]
    \centering
    \begin{minipage}[b]{\linewidth}
    \setlength{\tabcolsep}{0.8mm} 
    \small
    \begin{tabular}{@{\extracolsep{\fill}}c|lccc|c}
    \toprule
        \textbf{Methods} & \multicolumn{1}{c}{\textbf{Cat.}} & \textbf{SH} & \textbf{HS} & \textbf{FP} & \textbf{Average} \\ \midrule
        \multicolumn{6}{l}{\textbf{Llama-2}} \\ [3pt] 
        \multirow{2}{*}{P(True)} & Max & \underline{70.0} (0.4) & \underline{53.6} (1.1) & 76.1 (1.8) & 66.6 (1.1) \\
        ~ & Min & 68.2 (0.8) & 51.5 (1.2) & \underline{81.2} (1.1) & \underline{67.0} (1.0) \\
        \cmidrule{1-6}
        SelfCheck & Cer$_{S}$ & 66.0 (2.5) & 56.7 (1.2) & 77.3 (1.3) & 66.7 (1.7) \\
        GPT & Unc$_{S}$ & \underline{66.2} (1.7) & \underline{57.3} (1.9) & \underline{77.8} (1.1) & \underline{67.1} (1.6) \\
        \cmidrule{1-6}
        \multirow{3}{*}{Sampling} & Cer$_W$ & 67.9 (2.7) & 53.3 (2.1) & 78.8 (0.2) & 66.7 (1.7) \\
        ~ & Cer$_R$ & 66.9 (2.4) & \underline{53.8} (1.8) & \underline{80.0} (1.9) & 66.9 (2.0) \\
        ~ & Unc & \underline{69.1} (1.0) & \underline{53.8} (1.0) & 79.8 (1.2) & \underline{67.6} (1.1) \\
        \cmidrule{1-6}
        \multirow{3}{*}{\shortstack{Unc-TTP\\(ours)}} & Cer$_W$ & 68.5 (2.0) & \underline{\textbf{59.5}} (9.5) & 79.1 (1.7) & 69.0 (4.4) \\
        ~ & Cer$_R$ & 70.4 (1.9) & 56.9 (2.9) & 77.0 (3.4) & 68.1 (2.7) \\
        ~ & Unc & \underline{\textbf{71.2}} (2.1) & 58.8 (3.2) & \underline{\textbf{80.8}} (1.3) & \underline{\textbf{70.3}} (2.2) \\
        \midrule [1pt] 
        
        \multicolumn{6}{l}{\textbf{Mistral}} \\ [3pt] 
        \multirow{2}{*}{P(True)} & Max & 70.3 (0.6) & 80.7 (2.3) & \underline{81.5} (0.8) & \underline{77.5} (1.2) \\
        ~ & Min & \underline{73.2} (1.0) & \underline{80.7} (0.6) & 75.2 (0.8) & 76.4 (0.8) \\
        \cmidrule{1-6}
        SelfCheck & Cer$_{S}$ & \underline{70.2} (3.0) & \underline{\textbf{82.0}} (1.4) & 80.2 (2.9) & \underline{77.5} (2.4) \\
        GPT & Unc$_{S}$ & 65.5 (2.9) & 80.8 (2.1) & \underline{82.9} (1.5) & 76.4 (2.2) \\
        \cmidrule{1-6}
        \multirow{3}{*}{Sampling} & Cer$_W$ & 65.7 (5.1) & \underline{81.3} (1.8) & \underline{87.3} (1.9) & 78.1 (2.9) \\
        ~ & Cer$_R$ & 71.2 (3.8) & 77.3 (5.0) & 86.0 (1.4) & 78.2 (3.4) \\
        ~ & Unc & \underline{72.0} (0.9) & 79.5 (1.4) & 87.0 (1.2) & \underline{79.5} (1.2) \\
        \cmidrule{1-6}
         \multirow{3}{*}{\shortstack{Unc-TTP\\(ours)}} & Cer$_W$ & 71.2 (5.4) & 81.0 (2.3) & 86.4 (2.4) & 79.5 (3.4) \\
        ~ & Cer$_R$ & 66.6 (3.8) & 74.6 (2.0) & 84.3 (0.9) & 75.2 (2.2) \\
        ~ & Unc &  \underline{74.7} (0.8) & \underline{81.8} (2.9) & \underline{\textbf{88.1}} (0.8) & \underline{\textbf{81.5}} (1.5) \\
        \midrule [1pt]
        
        \multicolumn{6}{l}{\textbf{ GPT-3.5}} \\ [3pt]
        \multirow{2}{*}{P(True)} & Max & 74.8 (1.7) & 89.3 (0.9) & 80.2 (1.1) & 81.4 (1.2) \\
        ~ & Min & \underline{77.8} (1.0) & \underline{92.2} (0.6) & \underline{83.3} (2.2) & \underline{84.4} (1.3) \\
        \cmidrule{1-6}
        SelfCheck & Cer$_{S}$ & \underline{78.7} (2.9) & \underline{89.8} (0.5) & \underline{82.9} (0.6) & \underline{83.8} (1.3) \\
        GPT & Unc$_{S}$ & 77.3 (3.5) & 87.5 (0.7) & 81.0 (0.7) & 81.9 (1.6) \\
        \cmidrule{1-6}
        \multirow{3}{*}{Sampling} & Cer$_W$ & 78.5 (3.9) & 91.7 (0.8) & 83.0 (1.1) & 84.4 (1.9) \\
        ~ & Cer$_R$ & 77.3 (2.9) & 91.0 (1.3) & 82.7 (1.8) & 83.7 (2.0) \\
        ~ & Unc & \underline{80.1} (5.5) & \underline{92.5} (1.8) & \underline{83.5} (1.3) & \underline{\textbf{85.4}} (2.9) \\
        \cmidrule{1-6}
        \multirow{3}{*}{\shortstack{Unc-TTP\\(ours)}} & Cer$_W$ & 78.9 (3.6) & 88.2 (1.8) & 79.8 (1.4) & 82.3 (2.3) \\
        ~ & Cer$_R$ & \underline{82.8} (2.9) & 88.8 (2.4) & 82.1 (2.0) & 84.6 (2.4) \\
        ~ & Unc & 77.8 (2.4) & \underline{\textbf{93.0}} (0.5) & \underline{\textbf{84.8}} (1.7) & \underline{85.2} (1.5) \\
        \bottomrule
    \end{tabular}
    \caption{Complete results on three datasets for each of the four methods of measuring uncertainty through output inconsistency for each model.}
    \label{tab:main-unc-cer-test}
    \end{minipage}
\end{table}

\end{document}